*Review Article*

# A Systematic and Meta-Analysis Survey of Whale Optimization Algorithm

**Hardi M. Mohammed** ,[1,2] **Shahla U. Umar** ,[1,3] **and Tarik A. Rashid** [4]

[1]*Technical College of Informatics, Sulaimani Polytechnic University, Sulaimani, KRG, Iraq*
[2]*Applied Computer, College of Medicals and Applied Sciences, Charmo University, Sulaimani, Chamchamal, KRG, Iraq*
[3]*Network Department, College of Computer Science and Information Technology, Kirkuk University, Kirkuk, KRG, Iraq*
[4]*Computer Science and Engineering, University of Kurdistan Hewler (UKH), Erbil, KRG, Iraq*

Correspondence should be addressed to Hardi M. Mohammed; hardi.mohammed@charmouniversity.org

The whale optimization algorithm (WOA) is a nature-inspired metaheuristic optimization algorithm, which was proposed by Mirjalili and Lewis in 2016. This algorithm has shown its ability to solve many problems. Comprehensive surveys have been conducted about some other nature-inspired algorithms, such as ABC and PSO. Nonetheless, no survey search work has been conducted on WOA. Therefore, in this paper, a systematic and meta-analysis survey of WOA is conducted to help researchers to use it in different areas or hybridize it with other common algorithms. Thus, WOA is presented in depth in terms of algorithmic backgrounds, its characteristics, limitations, modifications, hybridizations, and applications. Next, WOA performances are presented to solve different problems. Then, the statistical results of WOA modifications and hybridizations are established and compared with the most common optimization algorithms and WOA. The survey's results indicate that WOA performs better than other common algorithms in terms of convergence speed and balancing between exploration and exploitation. WOA modifications and hybridizations also perform well compared to WOA. In addition, our investigation paves a way to present a new technique by hybridizing both WOA and BAT algorithms. The BAT algorithm is used for the exploration phase, whereas the WOA algorithm is used for the exploitation phase. Finally, statistical results obtained from WOA-BAT are very competitive and better than WOA in 16 benchmarks functions. WOA-BAT also outperforms well in 13 functions from CEC2005 and 7 functions from CEC2019.

## 1. Introduction

Recently, optimization becomes one of the most interesting issues in different life aspects, such as engineering designs, browsing the Internet, and business management. Time reduction, high quality, and financial profit can be challenging for the most real-world applications. Therefore, most optimization methods try to find a perfect method in order to deal with limited resources problem within various restrictions. Many effective search algorithms, which are using mathematical formulae and computational simulations, have been implemented to solve optimization problems. Metaheuristic algorithms try to balance between randomization and local search. So, most of these algorithms are used for global optimization [1, 2].

Metaheuristic algorithms have two basic elements, which are exploitation and exploration; in exploration, different solutions are found to explore the search space to find the global optimal, but in exploitation, local search is used by exploiting information about the best solutions that have been recently found. This combination with choosing the best solutions will guarantee that solutions reach the optimality, also exploration bypasses the local optima problem through randomization and raises the diversity of the solutions [1, 3].

Swarm-based nature metaheuristic algorithms are used to solve optimization problems by imitating the biological behavior of certain animals. Mirjalili and Lewis proposed the whale optimization algorithm to simulate the hunting behavior of humpback whales, and this is done by two main attacking mechanisms; first by chasing the prey with random or the best search agent and second by simulating the bubble net hunting strategy. Humpback whales like to hunt a group of small fish close to the surface. So, they swim around the target inside and alongside a thin circle to make a winding-shaped way, creating distinct blebs along a circle or '9' shaped ways altogether. Humpback whales have a very remarkable hunting method; this hunting behavior is called the bubble net feeding method. It has been observed that foraging is done by creating unique bubbles along a circle or '9' shaped path as shown in Figure 1 [5].

The aim of this research consists of several aspects: first of all, highlighting all studies and researches conducted on WOA, where metaheuristic hybridization models have been used to combine WOA with other techniques to enhance the performance of the resulting algorithm. Second, this work has focused on all modification methods, which have been applied on WOA to improve its ability to search for the best solution. Third, we have collected most of the research works related to various applications applied on WOA. Finally, a new hybridizing of WOA and BAT algorithms is presented. The proposed algorithm is used to overcome the problems of staying in the local optimum and increase the speed of convergence to the best solution. Consequently, this research work in return will pave the way for researchers to make other modifications on the WAO algorithm to suit their different purposes.

The rest of the paper outline starts with describing WOA, its characteristics, and limitations followed by providing various WOA modifications and hybridizations, which have been applied to different problems. Next, various applications of WOA are presented. After that, results from different benchmark functions and experiments are analyzed and compared to WOA modifications and other metaheuristic optimization algorithms. Then, the BAT algorithm is presented, and the WOA-BAT is proposed. The results of the WOA-BAT are evaluated against the original WOA. WOA-BAT is happened to be very competitive and better than WOA in 16 out of 23 benchmark test functions, 13 out of 25 CEC2005 test functions, and 7 out of 10 CEC2019 test functions. Finally, the conclusion is presented with future works on WOA and WOA-BAT.

### 1.1. Whale Optimization Algorithm.
This algorithm consists of two main phases; in the first phase, encircling prey and spiral updating position are implemented (exploitation phase). However, searching for a prey is done randomly in the second phase (exploration phase) [5]. The mathematical model of each phase is illustrated in the following subsections.

#### 1.1.1. Bubble Net Attacking Method.
Two approaches are designed in order to mathematically model the bubble net behavior of humpback whales that is called the exploitation phase. The two approaches are described as follows:

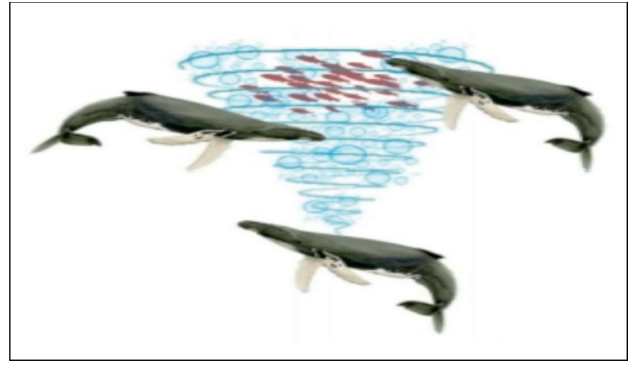

Figure 1: Spiral shape bubble net [4].

*(1) Encircling Prey.* After the humpback whales discover the position of the prey, they encircle around them. Firstly, the location of the optimal design in the search space is un-identified; thus, the WOA algorithm assumes that the present leading candidate solution is the target prey or near to the optimum. Then the other search agents will attempt to change their locations to the best search agents. This behavior is represented by the following equations:

$$\overrightarrow{X}(t+1) = \overrightarrow{X^*}(t) - \overrightarrow{A} \cdot \overrightarrow{D}, \tag{1}$$

$$\overrightarrow{D} = \left| \overrightarrow{C} \cdot \overrightarrow{X^*}(t) - \overrightarrow{X}(t) \right|, \tag{2}$$

where $\overrightarrow{X^*}(t)$ indicates the whale's earlier best location at iteration $t$. $\overrightarrow{X}(t+1)$ is the whale's current position, $\overrightarrow{D}$ is the distance vector between whale and prey, and $|\ |$ denotes absolute value. The $C$ and $A$ are coefficient vectors calculated as follows:

$$\overrightarrow{A} = 2 \cdot \overrightarrow{a} \cdot \overrightarrow{r} + \overrightarrow{a}, \tag{3}$$

$$\overrightarrow{C} = 2 \cdot \overrightarrow{r}. \tag{4}$$

To apply shrinking, the value of $\overrightarrow{a}$ is reduced in Equation (3); thus, the oscillating range of $\overrightarrow{A}$ is also reduced by $\overrightarrow{a}$. The value of $\overrightarrow{A}$ could be in the interval $(-a, a)$, where the value of $a$ is decreased from 2 to 0 through iterations. By selecting random values for $\overrightarrow{A}$ in $(-1, 1)$, the new position of any search agent can be determined anywhere in the range between the original position of the agent and the position of the current best agent.

*(2) Spiral Updating Position.* After calculating the distance between the whale located at $(X, Y)$ and prey located at $(X^*, Y^*)$. At that point, a spiral equation is generated between the location of the whale and prey to imitate the helix-shaped movement of humpback whales as follows:

$$\overrightarrow{X}(t+1) = e^{bk} \cdot \cos(2\pi k) \cdot \overrightarrow{D^*} + \overrightarrow{X^*}(t), \tag{5}$$

$$\overrightarrow{D^*} = \left| \overrightarrow{X^*}(t) - \overrightarrow{X}(t) \right|, \tag{6}$$

where $b$ is a constant value for identifying the shape of logarithmic spiral and $k$ is a random number in the range

[−1, 1]. This behavior is represented in WOA to change the position of whales during optimization. There is a 50% chance for selecting between the shrinking encircling mechanism and the spiral model, and their components are designed as follows:

$$\overrightarrow{X}(t+1) = \begin{cases} \overrightarrow{X^*} - \overrightarrow{A} \cdot \overrightarrow{D}, & \text{if } p < 0.5, \\ e^{bk} \cdot \cos(2\pi k) \cdot \overrightarrow{D^*} + \overrightarrow{X^*}(t), & \text{if } p \geq 0.5, \end{cases} \quad (7)$$

where $p$ is a random number in $(0, 1)$.

*1.1.2. Search for Prey.* In the search phase for the prey, which is called the exploration phase, a similar method depending on the variance of the $\overrightarrow{A}$ vector can be used. The whales actually use random search to discover their prey depending on the position of each other. Therefore, to oblige the search agents to move far away from the local whale, WOA uses the $\overrightarrow{A}$ vector with random values greater or less than 1. Throughout the exploration phase, the location of a search agent is reorganized according to randomly selected search agent rather than the best search agent (exploitation phase). This procedure aids the WOA algorithm to perform the global search and overcome the local optimal problem. The mathematical model is expressed as follows:

$$\overrightarrow{X}(t+1) = \overrightarrow{X_{\text{rand}}} - \overrightarrow{A} \cdot \overrightarrow{D}, \quad (8)$$

$$\overrightarrow{D} = \left| \overrightarrow{C} \cdot \overrightarrow{X_{\text{rand}}} - \overrightarrow{X} \right|, \quad (9)$$

where $\overrightarrow{X_{\text{rand}}}$ is the random position vector (a random whale) chosen from the current population.

*1.2. Operation of Whale Optimization Algorithm.* The WOA algorithm starts by assigning whales population with random solutions and assuming the best optimal value of the objective function is a minimum or maximum value (depending on the problem), then the objective function for each search agent is calculated. At each iteration, each search agent updates their location depending on either the best solution found so far when $|\overrightarrow{A}| < 1$ or on a randomly chosen search agent when $|\overrightarrow{A}| > 1$. In order to achieve exploration and exploitation phases, respectively, the value of $a$ parameter is decreased from 2 to 0. Also, WOA has the feature to select either a spiral or circular movement through the value of another parameter, which is $p$ (a random number in [0, 1]) with a probability of 50% to select one of these two mechanisms, so if its value is greater than 0.5, then the search agents change their positions using Equation (5), otherwise they use Equation (1). Finally, the WOA algorithm ends by implementing of the termination condition [5] (Algorithm 1).

*1.3. Whale Optimization Algorithm Pseudocode. 1.4. Characteristics of WOA.* The process of obtaining a suitable equivalence between exploitation and exploration in the improvement of any metaheuristic algorithm is a topmost challenge due to the arbitrary nature of the optimization algorithm. WOA has the highest significance compared to the different optimization approaches through the following:

(1) Exploitation ability
(2) Exploration ability
(3) Ability to get rid of the local minima

The WOA has an important capability of exploration due to the position updating mechanism of whales by using Equation (7). Throughout the initial step of the algorithm, this equation forces the whales to move randomly around each other. In the next steps, Equation (8) makes the whales update their positions rapidly and move along a spiral-shaped route in the direction of the best path that has been found so far. Since these two stages are done independently and in half iteration each, the WOA avoids local optima and achieves convergence speed at the same time through the iterations. But most of the other optimization algorithms (like PSO and GSA) do not have operators to consecrate a particular iteration to the exploration or the exploitation because they use only one format to update the position of search agents, so the probability of falling into local optima is more likely increased [2].

Figure 2 shows the number of publication that has been published about WOA since 2016.

*1.5. Limitation of WOA.* Metaheuristic algorithms have both efficiency and limitation for convergence speed and obtaining optimal solution. Thus, the limitation of WOA should also be found out depending on [6]. Randomization has a crucial role in exploration and exploitation, so using the current randomization technique in WOA would increase computational time especially for the highly complex problem [4].

Besides, convergence and speed depend on one control parameter, which is $a$. This parameter has an excessive impact on the performance of WOA [7]. For that reason, WOA has poor convergence speed in both exploration and exploitation phases [8, 9]. Thus, a balancing formulation between exploration and exploitation requires proper enhancement [10].

In addition, WOA uses the encircling mechanism in the search space, and this mechanism has less capability to jump out from local optima. Accordingly, it results in poor performance [11]. It also has a drawback when improving the best solution after each iteration [12].

It is worth mentioning that WOA cannot work in the fields of classification and dimensionality reduction as it is not suitable for the binary space [13]. Likewise, the original WOA cannot deal with complex environmental constraint as for the vehicle fuel consumption problem [14]. It cannot solve single and multidimensional 0–1 knapsack problems with different scales as it requires additional functions [15].

## 2. WOA Modifications and Hybridizations

In this review, we focus on reporting the developments of WOA, which have been published recently; this section is separated into three parts:

Details of the WOA are described below:
Generate initial population $X_i$ where ($i = 1, 2, 3, \ldots, n$)
Calculate the fitness of each solution
$X^*$ = the best search agent
   **While** ($t <$ Max_iterations)
     **For each solution**
       Update $a$, $A$, $C$, $L$, and $p$
       **If1** ($p < 0.5$)
         **If2** ($|A| < 1$)
           Update the position of the current search agent by using Equation (1)
         **Else if2** ($|A| \geq 1$)
           Select a random search agent ($X_{rand}$)
           Update the position of the current search agent by using Equation (7)
       **End if2**
      **Else if1** ($p \geq 0.5$)
         Update the position of the current search agent by using Equation (5)
      **End if1**
     **Enf for**
Check if any search agent goes beyond the search space and amend it
Calculate the fitness of each search agent
Update $X^*$ if there is a better solution $t = t + 1$
**End while**
Return $X^*$

ALGORITHM 1: The whale optimization algorithm pseudocode.

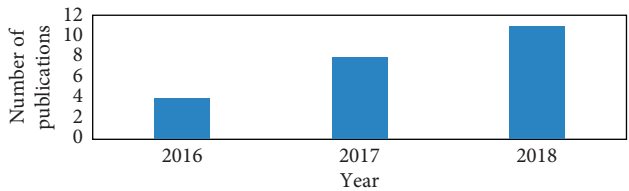

FIGURE 2: The number of publications on the whale optimization algorithm since 2016.

(a) Modifications of WOA: including AWOA, IWOA, chaotic WOA, ILWOA, and MWOA.

(b) Hybridizations of WOA: with metaheuristic algorithms, such as SA, PSO, local search, EWGC, and BS-WOA.

(c) Problems solved by WOA

### 2.1. Modifications of WOA.
There are different types of WOA, which have been modified. The following subsections are the summary of the modifications of WOA.

### 2.1.1. AWOA and SAWOA.
Randomizations have an essential influence in exploration and exploitation in optimization algorithms. Therefore, there are different techniques, which have been used in randomization, for example, Markov chain, Lévy flight, and normal distribution or Gaussian. Despite using these techniques, the adaptive technique has been used in WOA, which is called adaptive WOA (AWOA). This technique has also been used in the cuckoo search algorithm. This technique is crucial due to decreasing computational times for highly complicated problems [4, 16]. Having fewer parameters dependency is the best feature of this technique, which is useful, and does not need to initialize parameters and step sizes. Therefore, these parameters change regarding its fitness values during iterations. As a result, AWOA reached an optimum solution in less computational time and local optimum was avoided with the fast convergence [16, 17]. Trivedi et al. [4] proved that AWOA was better than WOA in terms of computational times and convergence speed.

Another AWOA was proposed in [18] for cluster head selection based on the Internet of things (IoT). IoT is another vital area, which can be researched on due to improvement in its performance [19, 20]. Regardless of using parameters, such as distance energy and delay of sensor nodes in a wireless sensor network, self-adaptiveness WOA (SAWOA) considers temperature and load parameters of IoT devices. Results proved that SAWOA performance was better than other algorithms like GSA, GA, ABC, PSO, and WOA [21–23].

### 2.1.2. IWOA.
Distance control parameter $a$ has value, which affects the ability of exploration and exploitation. However, this parameter is started from 2 and then decreased to 0 during iterations [7]. This parameter resulted in fast convergence and obtained accurate results for most problems. Despite these effects, it is linear and cannot adapt to the search process of WOA, which is nonlinear and complex [24]. Therefore, in an improved WOA (IWOA), some strategies defined for distance control parameter in order to adapt the nonlinear search process to achieve better results. There are five kinds of IWOA regarding its distance control variable, and those are SinWOA, CosWOA, TanWOA, LogWOA, and SquareWOA.

Because of having a poor balance between exploration and exploitation, researchers in [8] proposed a novel constitutional appraising approach based on WOA. Thus, results from clinical data analysis showed that IWOA had better efficiency in terms of convergence performance compared to the original WOA.

IWOA proposed in [25] used a new control parameter, which was inertia weight. This parameter was used to adjust the impact on the current best solution. To evaluate the performance of IWOA, 31 benchmark functions were used to test it in [5]. Then, IWOA was compared with WOA, FOA, and ABC algorithms. The size of the population was 1000 and with 30 iterations. IWOA outperformed compared to ABC, FOA, and WOA in terms of mean and standard deviations.

According to [26], the mean values of ABC and FOA were greater than IWOA and WOA for functions $f1$, $f2$, $f3$, $f7$, $f10$, $f11$, $f12$, $f13$, $f16$, and $f27$. The mean value of IWOA was the least compared to FOA and ABC for functions $f4'$, $f8'$, $f14$, $f15$, $f17$, $f18$, $f19$, $f20$, $f21$, and $f26$. However, the mean value for functions $f5$ and $f9$ was equal for all algorithms. The mean value of FOA was more than other algorithms for function $f6$. IWOA and FOA had greater mean values for functions $f23$, $f24$, and $f25$ compared to other algorithms.

The ABC mean value was better than IWOA, WOA, and FOA for functions $f''4$ and $f4$. Function $f22$ was unfit with all algorithms because the mean values were far away from the optimal value. The WOA mean value for $f''8$ had the least value. Similarly, the ABC mean value for $f8$ was the least. In addition, IWOA convergence was faster and obtained a lower value compared to WOA, FOA, and ABC. It can be said that IWOA was better than ABC and FOA. IWOA also enriched the original of WOA.

### 2.1.3. Chaotic WOA.

Metaheuristic algorithms have problems due to convergence speed and obtaining better performance. The theory of nonlinear chaos has widely been used in different applications [9]. Dynamical chaotic systems are able to control status and unsteady periodic motions. Chaos can be used in stochastic and deterministic algorithms. Due to developing the performance and convergence speed, chaos was used with WOA [27]. This theory has been used by variety of algorithms such as genetic algorithm [24], harmony search [28], PSO [29], ABC [30, 31], FA [32], KH [33], BOA [34], and GWO [35]. Different types of chaotic maps used with WOA in order to control the main parameter of WOA to provide stability between exploration and exploitation. Chaos means features of a complicated system, which have unpredictable behavior and map means relating parameters by using functions with chaos behavior. Ten unidimensional chaotic maps were used with WOA [36]. These ten maps were used to produce a chaotic set. The initial point was crucial because it had an impact on chaotic maps. As a result, 0.7 was chosen as an initial point, which ranges between 0 and 1 [37]. Therefore, 20 benchmark functions were tested by CWOA. As a result, chaotic maps enhanced the efficiency of WOA.

### 2.1.4. ILWOA.

Cloud computing is a computing system, which provides services via the Internet to clients [26]. Cloud computing is divided into two parts, which are the front end and back end. The front end includes all the software, which clients need, and the back end is related to the server and data storage [38]. Effectiveness and intelligent usages of cloud datacentre resources are the strategies of the consolidation of the virtual machine (VM). The most significant problem due to VM consolidation is VM replacement. Researchers aim was to minimize the number of physical machines, which were running in cloud datacentre. Abdel-Basset et al. [39] proposed improved Lévy WOA (ILWOA) to solve the minimization problem regarding the available bandwidth. Cloudsim toolkit was used to test the ILWOA on 25 different datasets and then compared with WOA, first fit, best fit, particle swarm optimization, genetic algorithm, and intelligent tuned harmony search. As a result, ILWOA showed better performance compared to other algorithms.

### 2.1.5. MWOA.

Because of developing technologies, protecting information is crucial in order to transmit it to the Internet. A modified version of WOA (MWOA), which was used for cryptanalysis of the Merkle–Hellman knapsack cryptosystem (MHKC), was developed in [40]. The continuous value was converted to discrete by a sigmoid function. Then, the evaluation function was dealt with an infeasible solution by adding a penalty function. Mutation operation was added to improve the solution. MHKC was the first public key cryptosystem (PKC) invented in 1987. Two keys were used by MHKC. The keys were private and public. Encrypting the plaintext was done by a public key, and decrypting was done by a private key [11]. MWOA was used to breakdown the MHKC by knowing ciphertext. Consequently, an attacker could reach the plaintext by using MWOA with the ciphertext [40].

### 2.1.6. Memetic WOA.

WOA is very competing with other common metaheuristic algorithms. However, WOA performance is restricted because of having search dynamics. Thus, the encircling mechanism mostly focuses on the exploration in the search space. As a result, WOA has poor performance to jump out from local optima. To solve this problem, memetic WOA (MWOA) was proposed in [12] by using chaotic local search inside WOA in order to extend the exploration capacity. MWOA was used to create stability between exploration and exploitation phases in the search space. To achieve a balance, MWOA was tested on 48 benchmark functions; then, results showed that MWOA performed better compared to its competitors with regard to accuracy and convergence speed.

### 2.2. Hybridization of WOA.

WOA was used with common metaheuristic algorithms to achieve better solutions and get rid of the weakness of WOA and other algorithms. The summaries of hybridizations of WOA are explained in the following subsections.

*2.2.1. With SA.* Majdi and Mirjalili [41] presented WOA with simulated annealing (SA). SA was embedded inside WOA to improve the best solution, which was found at the end of each iteration. WOA was able to search efficiently for finding the best solution. The blind operator was used by WOA in the exploitation phase, so this technique was replaced by using SA. Despite the effectiveness of WOA, SA was used to enhance the exploitation phase and overcome the stagnation in local optima.

*2.2.2. With PSO.* Trivedi et al. [42] used PSO and WOA to obtain a superior solution for global numerical functions. They used PSO for the exploitation phase, and WOA was used in exploration phase in an environment, which was not certain. WOA used the algorithmic spatial path to explore a possible solution in less computational time to avoid local optima [5]. The result showed the efficiency of PSO-WOA compared to PSO and WOA individually.

*2.2.3. With Local Search.* According to [40], the authors proposed WOA, a strategy that is called Local Search, in order to reduce permutation flow shop scheduling problem (FSSP). FSSP is an NP-hard problem, which is hard to find a result in polynomial time. Despite its essentiality, several algorithms have been developed to achieve two goals: reducing the time complexity and decreasing the duration of the best schedule. Other algorithms, which solved FSSP, had some drawbacks due to high computational cost and local optima [40]. Therefore, an algorithm was required for the largest rank value (LRV) to deal with the search space of the problem, which is discrete. As a result, a hybrid whale algorithm (HWA) was presented and able to achieve optimal solution quickly by using various techniques, for example, swap mutation, insert-reversed block operation, local search strategy, and integrated with a heuristic algorithm that is known as Nawaz–Enscore–Ham (NEH). Swap mutation operation was used to improve the diversity of the candidate schedule. Local optima were also avoided by using the insert-reversed block operation. As a result, HWA was combined with NEH to develop basic WOA performance. The proposed algorithm showed better results compared to the basic WOA [40].

*2.2.4. With EWGC.* Data are increasing nowadays; hence, controlling data becomes a difficult task. Therefore, data might be too complex. As a result, decision-making is affected by the way of organizing data. Thus, data clustering is essential to extract knowledge and make an efficient decision about knowledge. Exponential grey wolf optimization (EGWO) with whale optimization for data clustering (EWGC) was proposed to identify optimal centroid through the clustering process. WGC used the hybridization of WOA and WEGWO [43].

WGC used the WOA algorithm hunting mechanism to find centroid and position updates by using the EGWO algorithm in the exploration phase. Three datasets were used to test the proposed algorithm, and the results were compared with the particle swarm clustering (PSC), modified PS (mPS), grey wolf optimization (GWO), exponential GWO, kernel-based EGWO, and WOA. WGC showed better results compared to those algorithms.

*2.2.5. With BS.* Cloud computing has a major role in the digital era because it serves a large number of users at the same time. Besides of having many advantages, security of data, which are stored in the cloud platform, is a big challenge. Brainstorm WOA (BS-WOA) is a hybridized algorithm which is based on brainstorm optimization and WOA. Thus, BS-WOA was used to identify the secret key of the database because the privacy of users should be preserved. Therefore, BS-WOA generated a key for the data, which came from the data owner, in order to protect the data from being used by the third-party user. As a result, BS-WOA improved the privacy and utility of the data in the cloud while the secret key was identified during the optimization process [44].

# 3. Applications of WOA

WOA has been used in several areas in various academic and industrial fields so far and the most important application classes are shown in the following subsections.

*3.1. Electrical and Electronics Engineering.* In the last years, the distribution systems of electric power are requiring extensive voltage ratio to supply inductive loads, which cause more power losses in the distribution networks and weakness in the power factor. To control these problems, appropriate distribution of capacitors is provided and eliminating the network line losses could enhance the constancy and the accuracy of the system. In order to find the optimum sizing and status of the capacitors for standard radial distribution systems, WOA was proposed as a solution, and several aspects were taken into consideration, such as decreasing the cost of operating and minimizing the losses in the power with disparity limitation on the voltage range. The suggested algorithm was confirmed by applying it to standard radial systems: IEEE-34 and IEEE-85 bus radial distribution test systems. The obtained results were efficient compared with the existing algorithms [45].

The main function of the economic operation of power plants is scheduling the generating units to obtain minimum generation cost for the power utilities that means low-cost electricity. WOA is one of the most important new strategies to solve the economic dispatch problem. The execution of the utilized algorithm was verified using standard test system of IEEE 30-Bus; the obtained results from the proposed algorithm was compared with other metaheuristic approaches, such as PSO, ant colony optimization, and genetic algorithm and the comparison indicated that the obtained results were somewhat similar [46, 47].

*3.2. Economic Scheduling.* With a massive amount of real-world applications, the flow shop scheduling problem

(FSSP) has increased intensely. FSSP is regarded as an NP-hard problem since finding a solution in polynomial time is a difficult issue. In order to decrease the makespan of the best schedule and reduce the required time, WOA was merged with the local search technique for handling the flow shop scheduling problem. Swap mutation operation was utilized to enhance the diversity of item schedules, and the local optima problem was overcome by using insert-reverse block operation. The hybrid whale algorithm (HWA) obtained competitive results compared with the previous algorithms [10].

### 3.3. Civil Engineering.

The enhanced whale optimization algorithm (EWOA) was suggested to deal with sizing and optimization problems of truss and frame structures. EWOA was used to solve four structural optimization problems: two truss optimization problems (spatial 72-bar truss and spatial 582-bar tower) and two frame optimization problems (3-bay 15-story frame and 3-bay 24-story frame). The obtained numerical results showed that the suggested EWOA had better efficiency than the standard whale optimization algorithm [48].

### 3.4. Fuel and Energy.

WOA is widely used in the fields of saving, processing, and improving energy and fuel sources, and the following are some of these applications:

(1) The need for cleaning source of energy caused a rise in the using of solar energy; thus, researchers have given great importance to the design of photovoltaic cells. They faced two important problems; the first one was finding a beneficial model to describe the solar cells, and the second one was the lack of information about photovoltaic cells, which badly influences the efficiency of the photovoltaic modules (panels). The chaotic whale optimization algorithm (CWOA) for the parameter estimation of solar cells was made and used for calculating and automatically adapting the internal parameters of the optimization algorithm. The improved technique was able to optimize difficult and multimodal objective purpose. The experimental results of the proposed approach showed higher performance regarding accuracy and robustness [49].

(2) More recently, researchers have been searching for alternative energy sources, such as solar, wind, and biomass because of the lack of conventional energy sources, such as petrol and coal, and these sources are among the main causes of environmental pollution. At different circumstance and under variance conditions, it is very important to exploit the maximum solar power from the photovoltaic panels; thus, a modified artificial killer whale optimization algorithm (MAKWO) to trace and find the highest power region of the photovoltaic module in the partially cloudy atmosphere was suggested. The obtained findings from MAKWO were compared with different metaheuristic algorithms (modified wolf pack algorithm (MAWP), artificial bee colony (ABC), and particle swarm optimization (PSO)) with a significant performance for the proposed algorithm (MAWP) over all the other algorithms [50].

### 3.5. Medical Engineering.

Lately, analysis of medical images has become the focus of many researchers because they highly depend on these images for diagnosis and surgery. The liver is one of the organs most used in the computer-aided diagnosis system in order to detect the correct position of the organ inside the abdominal and also to avoid the intensity values overlapping with other organs. The whale optimization algorithm was proposed for liver segmentation in MIR images. To do the segmentation process, many clusters in the abdominal were determined. WOA had split the image into a number of clusters. After converting it to a binary image, it was multiplied by the previously clustered image with WOA in order to delete several parts of other organs; then, the required clusters were represented, which led to the liver area. A set of 70 images were tested using the suggested method illustrated and agreed by radiology specialists. Some measures like structural similarity index measure (SSIM), similarity index (SI), and other five measures were used to verify the correctness of the image. The final resolution of the processed image showed 96.75% accuracy using SSIM and 97.5 using SI% [51].

### 3.6. Problems Solved by WOA.

WOA is a metaheuristic optimization algorithm that can be used to solve different problems, such as engineering problems, binary problems, multiobjective problems, and scheduling problems. Table 1 summarizes several problems, which have been solved by the WOA.

## 4. Benchmark Functions Experiment

According to [21], WOA was compared with different algorithms, such as GSA, PSO, FEP, and DE. These algorithms were tested on 29 benchmark functions, it can be said that the benchmark functions are separated into four types: unimodal, multimodal, fixed-dimension multimodal, and composite functions, as shown in Table 2. These benchmark functions are used as a validation procedure to test WOA, and then the results are compared with other common algorithms to ensure whether WOA is better or not. Each algorithm runs 30 times in order to obtain the optimum solution.

The following subsections include comparison and discussion, solving classical engineering problem by WOA, comparison of WOA with IWOA, comparing WOA with other algorithms for feature selection, and finally, the evaluation performance of WOA is compared against AWOA and ILWOA.

### 4.1. Comparison and Discussion.

WOA characteristics were assessed based on 29 benchmark functions. These benchmark functions are standard functions that are used as a validation procedure to assess WOA and its modifications. The following sections have Tables 3–5, which show the average and standard deviation. The following points explain the exploitation, exploration, escaping from local minima, and convergence behavior.

TABLE 1: Problems solved by WOA.

| Method | Year, references | Problem | Purpose | Conclusion |
|---|---|---|---|---|
| WOA for constrained economic load dispatch problems | 2018, [13] | Economic load dispatch problem constraining | Giving reliable and constant electricity, whereas obtaining the best production with least cost and system operations | Solving the ELD problem resulted in the fast convergence and appropriate execution time |
| Binary WOA (bWOA) | 2018, [14] | Dimensionality reduction and classifications problem | Selecting the optimal feature subset, which can be the optimal solution based on the sigmoid transfer function (S shape) | bWOA could find optimal features, which have vital performance in terms of accuracy and execution time |
| Multiobjective method for vehicle traveling based on WOA (MOWOA) | 2017, [52] | Vehicle fuel consumption problem | Optimizing vehicle fuel consumption in terms of vehicle direction and traffic status | MOWOA satisfied the performance within the vehicle traveling optimization, and the performance increased slightly compared to Dijkstra's and A* algorithm |
| Using WOA | 2018, [53] | Nonuniformity in speed communication and illumination | Optimizing the position of the light emitting diodes (LEDs) | The result showed that this approach has given the higher uniformity compared to another result achieved by PSO |
| MOWOA | 2018, [54] | Multilevel threshold as a multiobjective function problem | Determining the multilevel threshold value for image segmentation | The result showed that WOA had better performance for solving this problem within faster convergence and lower execution time |
| Multiobjective task scheduling algorithm using WOA | 2017, [15] | The multiobjective task scheduling problem | Availability of low cost for each service and minimizing the execution time | The result showed great improvement in the proposed algorithm compared to original WOA |
| Improved whale optimization algorithm (IWOA) for solving both single and multidimensional problems | 2017, [55] | 0–1 knapsack problem | Handling infeasible solutions are the aim of this modification by adding penalty function to the evaluation function and sigmoid function to take the input parameter, which is the real-valued, and then produce the output | IWOA is able to give a balance between exploration and exploitation by using local search strategy (LSS) and the Lévy flight walks. The result indicated that IWOA is robust, effective, and efficient for solving this problem compared to other metaheuristic algorithms, which were used to solve this problem |
| The time-optimal memetic whale optimization algorithm | 2017, [56] | Hypersonic vehicle re-entry trajectory optimization problem with no-fly zones | Improving the robustness of IWOA to extend its strong ability on global search and improve the nonsensitivity of the initial values. Improve IWOA poor searching convergence speed by using Gauss pseudospectral methods (GPM) | Compared to the initial guess solution results of this hybridized technique, it concluded that it is very competitive and has better search accuracy, convergence speed, and robustness |

*4.1.1. Capability Exploitation Assessment.* F1 to F7 are unimodal functions that have only one local optimum. Therefore, by using them, we can evaluate the performance of exploitation of each algorithm. Table 3 shows that WOA is as good as other optimization algorithms for unimodal functions in exploration capability. Specifically, for F1 and F2, WOA is the most efficient optimizer and it has the second rank in almost all functions. Therefore, WOA is good at exploitation behavior [5].

TABLE 2: Description of unimodal, multimodal, fixed-dimension multimodal, and composite benchmark functions used in this work.

| Function | V_no | Range | $f_{min}$ |
|---|---|---|---|
| *Unimodal benchmark functions* | | | |
| 1　$f_1(x) = \sum_{i=1}^{n} x_i^2$ | 30 | [−100, 100] | 0 |
| 2　$f_2(x) = \sum_{i=1}^{n} |x_i| + \prod_{i=1}^{n} |x_i|$ | 30 | [−10, 10] | 0 |
| 3　$f_3(x) = \sum_{i=1}^{n} \left(\sum_{j-1}^{i} x_j\right)^2$ | 30 | [−100, 100] | 0 |
| 4　$f_4(x) = \max_i \{|x_i|, 1 \le i \le n\}$ | 30 | [−100, 100] | 0 |
| 5　$f_5(x) = \sum_{i=1}^{n-1} [100(x_{i+1} - x_i^2)^2 + (x_i - 1)^2]$ | 30 | [−30, 30] | 0 |
| 6　$f_6(x) = \sum_{i=1}^{n} ([x_i + 0.5])^2$ | 30 | [−100, 100] | 0 |
| 7　$f_7(x) = \sum_{i=1}^{n} i x_i^4 + \text{random}[0,1]$ | 30 | [−1.28, 1.28] | 0 |
| *Multimodal benchmark functions* | | | |
| 8　$f_8(x) = \sum_{i=1}^{n} -x_i \sin\left(\sqrt{|x_i|}\right)$ | 30 | [−500, 500] | −418.9829 × 5 |
| 9　$f_9(x) = \sum_{i=1}^{n} [x_i^2 - 10\cos(2\pi x_i) + 10]$ | 30 | [−5.12, 5.12] | 0 |
| 10　$f_{10}(x) = -20\exp\left(-0.2\sqrt{(1/n)\sum_{i=1}^{n} x_i^2}\right) - \exp\left((1/n)\sum_{i=1}^{n}\cos(2\pi x_i)\right) + 20 + e$ | 30 | [−32, 32] | 0 |
| 11　$f_{11}(x) = (1/4000)\sum_{i=1}^{n} x_i^2 - \prod_{i=1}^{n}\cos\left(x_i/\sqrt{i}\right) + 1$ | 30 | [−600, 600] | 0 |
| 12　$f_{12}(x) = (\pi/n)\left\{10\sin(\pi y_1) + \sum_{i=1}^{n-1}(y_i-1)^2[1 + 10\sin^2(\pi y_{i+1})] + (y_n-1)^2\right\} + \sum_{i=1}^{n}\mu(x_i, 10, 100, 4)$ | 30 | [−50, 50] | 0 |
| $y_i = 1 + (x_i + 1/4)\mu(x_i, a, k, m) = \begin{cases} k(x_i - a)^m & x_i > a \\ 0 & -a < x_i < a \\ k(-x_i - a)^m & x_i < -a \end{cases}$ | | | |
| 13　$f_{13}(x) = 0.1\left\{\sin^2(3\pi x_1) + \sum_{i=1}^{n}(x_i - 1)^2[1 + \sin^2(3\pi x_i + 1)] + (x_n - 1)^2[1 + \sin^2(2\pi x_n)]\right\}$ $+ \sum_{i=1}^{n}\mu(x_i, 10, 100, 4)$ | 30 | [−50, 50] | 0 |
| *Fixed-dimension multimodal benchmark functions* | | | |
| 14　$f_{14}(x) = \left((1/500) + \sum_{j=1}^{25} \left(1/\left(j + \sum_{i=1}^{2}(x_i - a_{ij})^6\right)\right)\right)^{-1}$ | 2 | [−65, 65] | 1 |
| 15　$f_{15}(x) = \sum_{i=1}^{11} [a_i - (x_1(b_i^2 + b_i x_2)/(b_i^2 + b_i x_3 + x_4)]^2$ | 4 | [−5, 5] | 0.00030 |
| 16　$f_{16}(x) = 4x_1^2 - 2.1x_1^4 + (1/3)x_1^6 + x_1 x_2 - 4x_2^2 + 4x_2^4$ | 2 | [−5, 5] | −1.398 |
| 17　$f_{17}(x) = (x_2 - (5.1/4\pi^2)x_1^2 + (5/\pi)x_1 - 6)^2 + 10(1 - (1/8\pi))\cos x_1 + 10$ | 2 | [−5, 5] | 0.398 |
| 18　$f_{18}(x) = [1 + (x_1 + x_2 + 1)^2 (19 - 14x_1 + 3x_1^2 - 14x_2 + 6x_1x_2 + 3x_2^2)] \times [30 + (2x_1 - 3x_2)^2$ $\times (18 - 32x_1 + 12x_1^2 + 48x_2 - 36x_1x_2 + 27x_2^2)]$ | 2 | [−2, 2] | 3 |
| 19　$f_{19}(x) = -\sum_{i=1}^{4} c_i \exp\left(-\sum_{j=1}^{3} a_{ij}(x_j - p_{ij})^2\right)$ | 3 | [1, 3] | −3.86 |
| 20　$f_{20}(x) = -\sum_{i=1}^{4} c_i \exp\left(-\sum_{j=1}^{6} a_{ij}(x_j - p_{ij})^2\right)$ | 6 | [0, 1] | −3.32 |
| 21　$f_{21}(x) = -\sum_{i=1}^{5} [(x - a_i)(x - a_i)^T + c_i]^{-1}$ | 4 | [0, 10] | −10.1532 |
| 22　$f_{22}(x) = -\sum_{i=1}^{7} [(x - a_i)(x - a_i)^T + c_i]^{-1}$ | 4 | [0, 10] | −10.4028 |
| 23　$f_{23}(x) = -\sum_{i=1}^{10} [(x - a_i)(x - a_i)^T + c_i]^{-1}$ | 4 | [0, 10] | −10.5363 |

Table 3: Result comparison among optimization algorithms [2].

| F | DE | | GSA | | PSO | | FEP | | WOA | |
|---|---|---|---|---|---|---|---|---|---|---|
| | avg | std | avg | std | avg | std | avg | std | avg | std |
| F1 | $8.2E - 14$ | $5.9E - 14$ | $2.53E - 16$ | $9.67E - 17$ | 0.000136 | 0.000202 | 0.00057 | 0.00013 | $1.41E - 30$ | $4.91E - 30$ |
| F2 | $8.2E - 14$ | $5.9E - 14$ | $2.53E - 16$ | $9.67E - 17$ | 0.000136 | 0.000202 | 0.00057 | 0.00013 | $1.41E - 30$ | $4.91E - 30$ |
| F3 | $1.5E - 09$ | $9.9E - 10$ | 0.055655 | 0.194074 | 0.042144 | 0.045421 | 0.0081 | 0.00077 | $1.06E - 21$ | $2.39E - 21$ |
| F4 | $6.8E - 11$ | $7.4E - 11$ | 896.5347 | 318.9559 | 70.12562 | 22.11924 | 0.016 | 0.014 | $5.39E - 07$ | $2.93E - 06$ |
| F5 | 0 | 0 | 7.35487 | 1.741452 | 1.086481 | 0.317039 | 0.3 | 0.5 | 0.072581 | 0.39747 |
| F6 | 0 | 0 | 67.54309 | 62.22534 | 96.71832 | 60.11559 | 5.06 | 5.87 | 27.86558 | 0.763626 |
| F7 | 0 | 0 | $2.5E - 16$ | $1.74E - 16$ | 0.000102 | $8.28E - 05$ | 0 | 0 | 3.116266 | 0.532429 |

*4.1.2. Capability Exploration Assessment.* Multimodal functions have various local optima, while unimodal has one local best. Therefore, the local optimal number increases when the number of design variables increases. As a result, these types of function are vital to evaluate the exploration capability over other optimal algorithms. Table 4 shows that WOA has a good capability for exploration. Because of the integration mechanism of exploration, WOA has the second rank compared with other optimization algorithms.

*4.1.3. Escaping from Local Minima.* Balancing between exploration and exploitation is the only way to avoid local optima because of challenging of mathematical computation of a composite function. Table 5 shows that the WOA algorithm ranked as the first optimizer in three tests and is as good as other optimization algorithms. It also demonstrates that WOA works well to make a balance between exploration and exploitation phases.

*4.1.4. Convergence Behavior Analysis.* When comparing different metaheuristic algorithms (WOA, PSO, and GSA) for some problems, it can be seen that the convergence rate of WOA is well competitive with other algorithms when it is tested on 29 benchmarks functions [5]. WOA has many main characteristics that make it faster than other algorithms. In the initial steps of iterations, the search agents try to relocate their positions randomly around each other through Equation (8), which gives WOA high exploration capability, while using Equation (7), the search agents reposition their locations in a spiral-shaped path toward the best solution found so far. Each phase is done in almost half of iterations and simultaneously; thus, the WOA has the highest local optima avoidance capability and fast convergence rate than other similar metaheuristic algorithms. However, PSO and GSA have a greater probability of falling into starvation in local optima simply because they do not have parameters to determine specific iterations to the exploration or exploitation phases. In other words, they use only one equation to update the search agents' positions, and also WOA requires less iteration to obtain global optimum compare to other algorithms.

*4.2. WOA for Classical Engineering Problem.* Mirjalili and Lewis [5] used WOA to solve the following engineering problems, which are shown in Table 6.

*4.3. WOA Feature Selection Experiment.* 16 datasets were chosen in this paper [28]. Training, validation, and testing were steps in which datasets were used. Each dataset randomly separated into three parts. The classification was done by the training part, and the validation part was used to assess the classification capability. Finally, the test part was required for evaluating the selected features. WOA, PSO, and GA were used on this test in order to achieve the comparison results.

The results were computed in the Matlab environment 20 times. Overall, WOA outperformed feature selections, which approved the ability of wrapper-based approach and premature convergence while searching for optimal feature subset in the search space. WOA was better than PSO and GA in terms of ability to search for optimal features. Occurring local optima that may happen because of premature convergence can be avoided by WOA. Moreover, the results proved that WOA can find an optimal solution, which had maximum classification accuracy. It was also capable to make stability between exploration and exploitation.

*4.4. Performance Evaluation on Benchmark Functions between Several Variants of WOA.* WOA has different types of modification. Therefore, it was compared with AWOA and ILWOA in the following subsections.

*4.4.1. WOA and AWOA Comparison.* AWOA was tested on different unconstraint benchmark functions. It can be said that AWOA had a better result compared to WOA. AWOA improved solution by fast convergence, randomness, and stochastic behavior. It was also used as a random search in workspace while no optimal solutions exist. Thus, AWOA was an effective technique to solve the problem within unknown search space [4].

*4.4.2. WOA and ILWOA.* Statistical results from [12] show the difference between ILWOA and WOA performance, Friedman test is used with the experimental result to analysis. Friedman test can be executed on more than two dependent samples because it is a non-parametric and rank-based version of one-way ANOVA with respected measures.

Table 4: Result comparison among optimization algorithms [5].

| F | DE | | GSA | | PSO | | FEP | | WOA | |
|---|---|---|---|---|---|---|---|---|---|---|
| | avg | std | avg | std | avg | std | avg | std | avg | std |
| F8 | −11080.1 | 574.7 | −2821.07 | 493.0375 | −4841.29 | 1152.814 | −12554.5 | 52.6 | −5080.76 | 695.7968 |
| F9 | 69.2 | 38.8 | 25.96841 | 7.470068 | 46.70423 | 11.62938 | 0.046 | 0.012 | 0 | 0 |
| F10 | 7.4043 | 4.2E − 08 | 0.06207 | 0.23628 | 0.27605 | 0.50901 | 0.018 | 0.0021 | 7.4043 | 9.897572 |
| F11 | 0.000289 | 0 | 27.70154 | 5.040343 | 0.009215 | 0.007724 | 0.016 | 0.022 | 0.000289 | 0.001586 |
| F12 | 0.339676 | 8E − 15 | 1.799617 | 0.95114 | 0.006917 | 0.026301 | 9.2E − 06 | 3.6E − 06 | 0.339676 | 0.214864 |
| F13 | 1.889015 | 4.8E − 14 | 8.899084 | 7.126241 | 0.006675 | 0.008907 | 0.00016 | 0.000073 | 1.889015 | 0.266088 |
| F14 | 2.111973 | 3.3E − 16 | 5.859838 | 3.831299 | 3.627168 | 2.560828 | 1.22 | 0.56 | 2.111973 | 2.498594 |
| F15 | 0.000572 | 0.00033 | 0.003673 | 0.001647 | 0.000577 | 0.000222 | 0.0005 | 0.00032 | 0.000572 | 0.000324 |
| F16 | −1.03163 | 3.1E − 13 | −1.03163 | 4.88E − 16 | −1.03163 | 6.25E − 16 | −1.03 | 4.9E − 07 | −1.03163 | 4.2E − 07 |
| F17 | 0.397887 | 9.9E − 09 | 0.397887 | 0 | 0.397887 | 0 | 0.398 | 1.5E − 07 | 0.397914 | 2.7E − 05 |
| F18 | 3 | 2E − 15 | 3 | 4.17E − 15 | 3 | 1.33E − 15 | 3.02 | 0.11 | 3 | 4.22E − 15 |
| F19 | N/A | N/A | −3.86278 | 2.29E − 15 | −3.86278 | 2.58E − 15 | −3.86 | 0.000014 | −3.85616 | 0.002706 |
| F20 | N/A | N/A | −3.31778 | 0.023081 | −3.26634 | 0.060516 | −3.27 | 0.059 | −2.98105 | 0.376653 |
| F21 | −10.1532 | 0.0000025 | −5.95512 | 3.737079 | −6.8651 | 3.019644 | −5.52 | 1.59 | −7.04918 | 3.629551 |
| F22 | −10.4029 | 3.9E − 07 | −9.68447 | 2.014088 | −8.45653 | 3.087094 | −5.53 | 2.12 | −8.18178 | 3.829202 |
| F23 | −10.5364 | 1.9E − 07 | −10.5364 | 2.6E − 15 | −9.95291 | 1.782786 | −6.57 | 3.14 | −9.34238 | 2.414737 |

Table 5: Composite benchmark functions comparison result [5].

| F | DE | | GSA | | PSO | | WOA | |
|---|---|---|---|---|---|---|---|---|
| | avg | std | avg | std | avg | std | avg | std |
| F24 | 6.75E − 2 | 6.75E − 2 | 6.75E − 2 | 2.78E − 17 | 100 | 81.65 | 0.568846 | 0.505946 |
| F25 | 28.759 | 8.6277 | 200.6202 | 67.72087 | 155.91 | 13.176 | 75.30874 | 43.07855 |
| F26 | 144.41 | 19.401 | 180 | 91.89366 | 172.03 | 32.769 | 55.65147 | 21.87944 |
| F27 | 324.86 | 14.784 | 170 | 82.32726 | 314.3 | 20.066 | 53.83778 | 21.621 |
| F28 | 10.789 | 2.604 | 200 | 47.14045 | 83.45 | 101.11 | 77.8064 | 52.02346 |
| F29 | 490.94 | 39.461 | 142.0906 | 88.87141 | 861.42 | 125.81 | 57.88445 | 34.44601 |

Table 6: Different engineering problem comparison result.

| Problems | Aim | Result |
|---|---|---|
| Tension/ compression spring design problem | Minimizing the weight of tension/ compression spring is the goal of this design problem | WOA had better performance over PSO and GSA on average, and both PSO and GSA required more function evaluation than WOA [5] |
| Welded beam design problem | Minimizing the fabrication cost of the welded beam is the objective | WOA outperformed over PSO and GSA on average and required the least number of function evaluations to find the best optimal solution |
| Pressure vessel design | The objective is to minimize the total cost of a cylindrical vessel | WOA performed better compared to PSO and GSA on average and the required of a number of evaluation function [5] |
| 15-bar truss design problem | Minimizing the weight of the 15-bar truss is the goal of this problem | WOA had similar performance, which would find a similar structure with other algorithms. WOA had the second rank for the number of the evaluation function |

Figures 3 and 4 show datasets with three and five types of hosts with Friedman ranked mean for each algorithm. The result showed that ILWOA had the best performance in minimizing utilization of host machines [39]. Friedman test and datacentre utilization host were performed to analyze the obtaining result. It is clear that ILWOA had been tested on three and five bin datasets. As a result, the efficiency of ILWOA increased as the number of bin datasets increased.

## 5. Standard Bat Algorithm

The bat algorithm is a metaheuristic algorithm developed by XinShe Yang in 2010 [57]. It was based on the echolocation capabilities of microbats. Before illustrating the details of this algorithm, we summarize echolocation briefly.

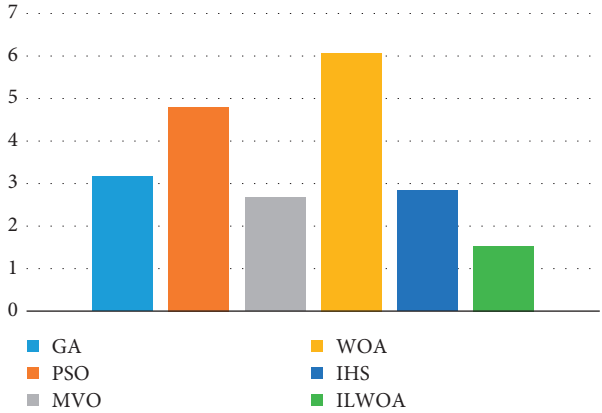

Figure 3: Friedman test of datasets with 3 host types.

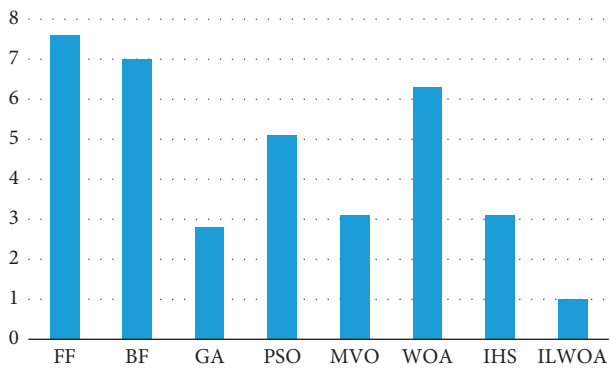

Figure 4: Friedman test of datasets with 5 host types.

*5.1. Echolocation of Microbats.* Bats are mammals with echolocation capabilities. They use echolocation sonar to detect prey or to avoid obstacles. These bats send a very loud sound pulse and receive the echo that rebounds back from the surrounding objects. In zones with identical atmospheric air pressure, these sound pluses emit at a constant velocity while they get changed if the atmospheric pressure is changed [57]. Bats can estimate the positions of any surrounding objects using the time delay of the returning pulse, and also they determine the shape and the direction of the objects using comparative amplitudes of the sound pulses collected at each ear. Finally, the data obtained so far are analyzed and interpreted in the brain to construct image about their surroundings [58].

*5.2. Bat Algorithm.* Using the concept of bat echolocation abilities, Yang (2010) developed various bat-inspired algorithms or bat algorithms. He simulated this behavior to solve different optimization problems. Bats can determine the position of their preys, objects, or food exactly through very loud sound wave emission and receiving echo that comes back from these objects. Bats use the advantages of time delay concept to find their preys, whereas the time delay is calculated as space between bats' ears and the echo wave variations. On a trip finding the prey, bats fly randomly in the search space with a speed $v_i$ and change their positions $x_i$ at a constant frequency $f_{min}$, different wavelengths $\beta$, and loudness $A_0$. In the algorithm, to update the value of these parameters, Yang used the following three equations [57]:

$$f_i = f_{min} + (f_{max} - f_{min})\beta, \tag{10}$$

$$v_i^{t+1} = v_i^t + (x_i^t - x_*)f_i, \tag{11}$$

$$x_i^{t+1} = x_i^t + v_i^t, \tag{12}$$

where $x_i$ is the position of the bats, $v_i$ is the velocity of bats, $f_i$ is the frequency of waves, and $\beta$ is a random vector in the range [0, 1] taken from regular distribution. However, $x_*$ refers to the global best solution found so far among all bats in the search space. Based on the domain size of the optimization problem, the upper and the lower limits of the frequency are determined. Usually, the upper boundary is assigned 100 and the lower boundary is assigned 0. Initially, each bat takes a random value of the frequency within the range $(f_{min}, f_{max})$. The velocity of the search agent is compatible with the frequency, and the position of the new solution is located depending on its new velocity. When a bat finds its prey or food, the rate of the loudness reduces while the ratio of pulse emission rises. A pseudocode listed by Yang (2010) is shown in Algorithm 2.

## 6. Hybrid WOA-BAT Algorithm

WOA is an optimization algorithm, which showed high performance in solving many optimization problems. Despite all the results, WOA showed slow convergence speed due to finding the global optimum [53]. Therefore, the BAT algorithm is used to improve the exploration capability of WOA. In this approach, two basic techniques are used: (1) the BAT algorithm is partially embedded inside WOA search phase and (2) the condition technique is used after changing the position of each search agent; for example, if the new position is better than the old position, then the old position is replaced. As a result, the WOA-BAT algorithm can obtain better results in fewer iterations compared to WOA. The detail of this modification can be seen in the WOA-BAT algorithm pseudocode (Algorithm 3) and Figure 5.

## 7. Implementation and Results

The proposed algorithm WOA-BAT is implemented and evaluated by using different benchmark functions. Three different benchmark functions are used to test the proposed algorithm; these are 23 mathematical optimization problems (Table 2 and [5]), CEC2005 (Table 7), and CEC2019 (Table 8). The code of WOA-BAT is available at the following link: https://github.com/Hardi-Mohammed/WOA-BAT-modification. In order to improve the WOA code which have been implemented by Mirjalili and Lewis, WOA and BAT algorithms are hybridized using the Matlab code. The following subsections include a description of benchmark

```
Objective function  f (x), x = (x₁, . . . , x_d)ᵀ
Initialize the bat individuals  xᵢ, (i = 1, 2, . . . , n) and vᵢ
Set pulse frequency  fᵢ at xᵢ
Initialize pulse rates rᵢ and the loudness  Aᵢ
While (t < maximum number of iterations)
Generate new solution by adjusting frequency and updating velocities and locations/solutions (Equations (2)–(4))
      if (rand > rᵢ)
      Select a solution among the best solutions
      Generate a local solution around the selected best solution
      End if
      Generate a new solution by flying randomly
      If (rand < Aᵢ & f (xᵢ) < f (x∗))
      Accept the new solutions
      Increase rᵢ and decrease Aᵢ
      End if
Rank the bats and find the current best x∗
End while
```

ALGORITHM 2: BAT algorithm pseudocode.

```
Details of the WOA are described below:
Generate initial population Xi where (i = 1, 2, 3, . . ., n)
Initialize f, v, r, and A1
Initialize fMin, fMax
Calculate the fitness of each solution
X∗ = the best search agent
   While (t < Max_iterations)
      For each solution
         Update a, A, C, L, and p
            If1 (p < 0.5)
               If2 (/A/ < 1)
                  Update the position of the current search agent by using Equations (10)–(12)
               Else if2 (/A/ ≥ 1)
                  Select a random search agent (Xrand)
                  Update the position of the current search agent by using Equations (10)–(12)
               End if2
            Else if1 (p ≥ 0.5)
               Update the position of the current search agent by using Equation (5)
            End if1
         Enf for
   Check if any search agent goes beyond the search space and amend it
   Calculate the fitness of each search agent
   Update X∗ if there is a better solution t = t + 1
   End while
   Return X∗
```

ALGORITHM 3: WOA-BAT algorithm pseudocode.

functions, experimental setup, evaluation criteria, comparison of WOA-BAT with WOA, and comparison of WOA-BAT with common algorithms.

### 7.1. Benchmark Functions.

First, the implementation of 23 mathematical benchmark functions is conducted [5]. The test functions can be classified into two groups: $f1$–$f7$ (unimodal benchmark functions) and $f8$–$f23$ (multimodal benchmark functions). Second, CEC2005 includes four types of benchmark functions; these are unimodal functions, multimodal functions, expanded multimodal functions, and hybrid composition functions [59–61]. CEC2005 function details are in Table 7 [62]. Each part includes the numbers, respectively, 5, 7, 2, and 11. Third, 10 benchmark functions are used from CEC2019. All the benchmark functions in CEC2019 are multimodal functions and can be seen in Table 8 [63].

### 7.2. Experimental Setup.

To obtain an accurate result, population size is randomly generated in order to make the best comparison with other common algorithms. The population size is 30, which were randomly generated, maximum iteration for the population size is 500, and the dimension is 30. The population size and iterations are executed 30 times then the average result is taken.

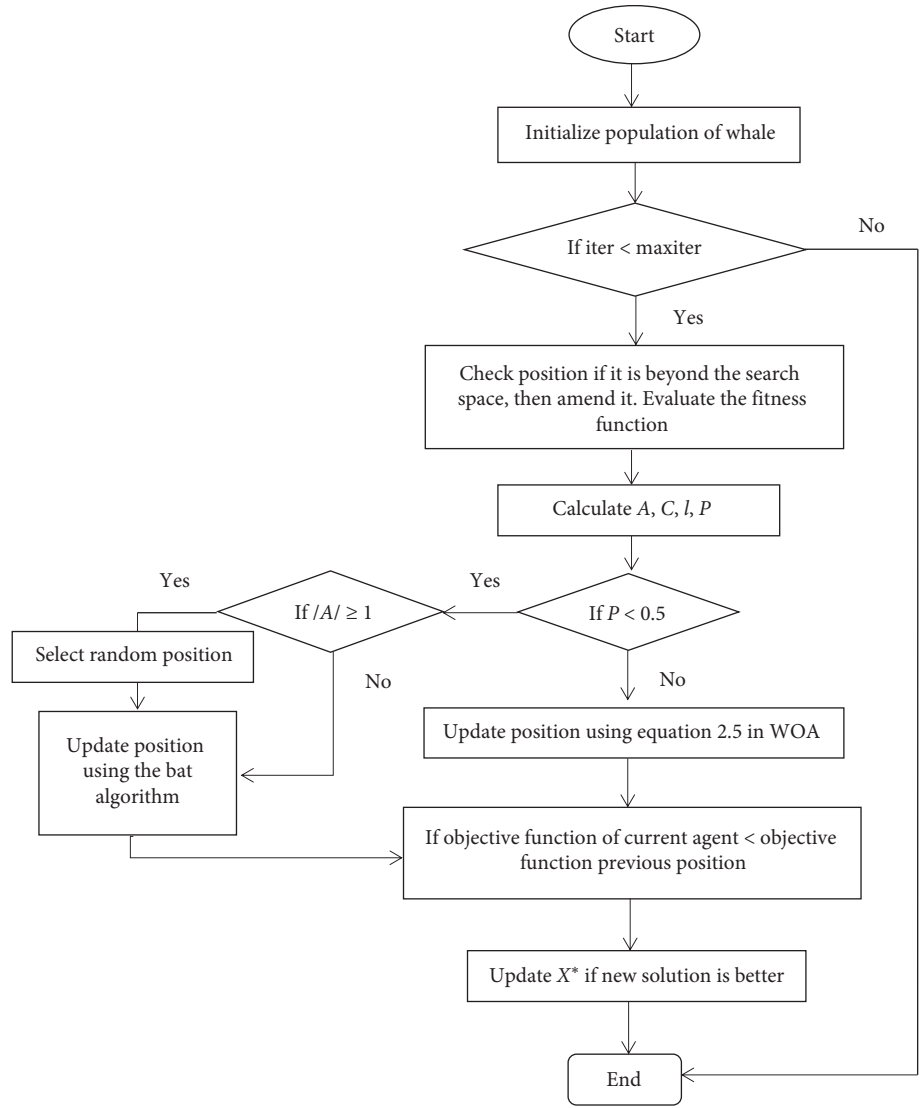

FIGURE 5: WOA-BAT flowchart.

### 7.3. Evaluation Criteria.
Three ways are used to evaluate algorithms to obtain better comparison, and the following points are the criteria of evaluation:

(1) Obtaining average and standard deviation

(2) Comparing WOA-BAT by building a box and whisker plot with WOA

(3) Compare WOA-BAT with other metaheuristic algorithms

### 7.4. Comparison with Original WOA Algorithm

#### 7.4.1. Evaluation of F1–F7 Exploitation.
These are unimodal functions as they have a single optimum global value. By using these functions, we can easily investigate the exploitation capability of the developed algorithm. Therefore, Table 9 and Figure 6 show that WOA-BAT as a better optimizer for $f3$, $f4$, $f5$, $f6$, and $f7$, while WOA is better for $f1$ and $f2$. As a result, WOA-BAT has an effective ability in exploitation.

#### 7.4.2. Evaluation of F8–F23 Exploration.
Functions $f8$–$f23$ are multimodal functions, which can have many local optima. These numbers of local optima are increased depending on the design variables. Therefore, these functions can be used to test the exploration capability of the WOA-BAT algorithm. Table 9 and Figure 6 illustrate that the averages of 10 benchmark functions ($f8$, $f11$, $f12$, $f13$, $f14$, $f15$, $f17$, $f19$, $f20$, $f22$, and $f23$) are very efficient in WOA-BAT, while WOA has the optimum value in 4 functions ($f9$, $f10$, $f18$, and $f21$).

25 benchmark functions of CEC2005 are used to test on WOA-BAT and WOA. Table 10 and Figure 7 show the comparison results of WOA and WOA-BAT in box and whisker plot. WOA-BAT outperforms well in 13 functions.

Table 7: Summary of the 25 CEC2005 test functions.

| No. | Functions | $F_1(x^*) = f\_bias_i$ | $D$ | Search range |
|---|---|---|---|---|
| *Unimodal benchmark functions* (5) | | | | |
| 1 | Shifted sphere function | −450 | 10, 30, 50 | [−100, 100] |
| 2 | Shifted Schwefel's problem 1.2 | −450 | 10, 30, 50 | [−100, 100] |
| 3 | Shifted rotated high conditioned elliptic function | −450 | 10, 30, 50 | [−100, 100] |
| 4 | Shifted Schwefel's problem 1.2 with noise in fitness | −450 | 10, 30, 50 | [−100, 100] |
| 5 | Schwefel's problem 2.6 with global optimum on bounds | −310 | 10, 30, 50 | [−100, 100] |
| *Multimodal functions Basic functions* (7) | | | | |
| 6 | Shifted Rosenbrock's function | 390 | 10, 30, 50 | [−100, 100] |
| 7 | Shifted rotated Griewank function without bounds | −180 | 10, 30, 50 | [0, 600] |
| 8 | Shifted rotated Ackley's function with global optimum on bounds | −140 | 10, 30, 50 | [−32, 32] |
| 9 | Shifted Rastrigin's function | −330 | 10, 30, 50 | [−5, 5] |
| 10 | Shifted rotated Rastrigin's function | −330 | 10, 30, 50 | [−5, 5] |
| 11 | Shifted rotated weierstrass function | 90 | 10, 30, 50 | [−0.5, 0.5] |
| 12 | Schwefel's problem 2.13 | −460 | 10, 30, 50 | [−π, π] |
| *Expanded functions* (2) | | | | |
| 13 | Expanded extended Griewank plus Rosenbrock's function (F8F2) | −130 | 10, 30, 50 | [−3, 1] |
| 14 | Shifted rotated expanded Scaffer's F6 | −300 | 10, 30, 50 | [−100, 100] |
| *Hybrid composition functions* (11) | | | | |
| 15 | Hybrid composition function | 120 | 10, 30, 50 | [−5, 5] |
| 16 | Rotated hybrid composition function | 120 | 10, 30, 50 | [−5, 5] |
| 17 | Rotated hybrid composition function with noise in fitness | 120 | 10, 30, 50 | [−5, 5] |
| 18 | Rotated hybrid composition function | 10 | 10, 30, 50 | [−5, 5] |
| 19 | Rotated hybrid composition function with a narrow basin for the global optimum | 10 | 10, 30, 50 | [−5, 5] |
| 20 | Rotated hybrid composition function with the global optimum on the bounds | 10 | 10, 30, 50 | [−5, 5] |
| 21 | Rotated hybrid composition function | 360 | 10, 30, 50 | [−5, 5] |
| 22 | Rotated hybrid composition function with high condition number matrix | 360 | 10, 30, 50 | [−5, 5] |
| 23 | Noncontinuous rotated hybrid composition function | 360 | 10, 30, 50 | [−5, 5] |
| 24 | Rotated hybrid composition function | 260 | 10, 30, 50 | [−5, 5] |
| 25 | Rotated hybrid composition function without bounds | 260 | 10, 30, 50 | [2, 5] |

Table 10 shows that WOA-BAT has better performance compared to WOA original in $f1$, $f2$, $f3$, $f4$, $f6$, $f9$, $f10$, $f12$, $f13$, $f18$, $f19$, $f22$, and $f25$. However, WOA outperforms in other functions while WOA-BAT and WOA have the same performance in $f7$ and $f8$, which can be seen in Figure 7. Overall, it can be said that the proposed algorithm improved the WOA original to obtain a better result in approximately 13 functions.

Like CEC2005, CEC2019 is used to test the WOA-BAT algorithm and WOA. Table 11 and Figure 8 show that WOA-BAT has lower average result compared to WOA in eight functions $f1$, $f2$, $f3$, $f5$, $f7$, $f8$, and $f10$. However, WOA-BAT is not very competitive with WOA in $f4$, $f6$, and $f9$. Overall, WOA-BAT could improve the WOA in 7 benchmark functions from CEC2019.

*7.5. Comparison with Metaheuristic Algorithms.* The results from different papers included and presented in this paper in order to compare WOA-BAT with other well-known evolutionary algorithms, for example, GA, DE, ABC, and BSO.

The results of these algorithms are obtained from CEC2005, which includes 25 benchmark functions [59–61]. The results for CEC2005 as shown in Table 12 indicate that WOA-BAT has the first rank because it outperforms well in 13 functions. The function, which WOA-BAT has a better result, are $f3$, $f11$, $f12$, $f15$, $f16$, $f17$, $f18$, $f19$, $f20$, $f21$, $f22$, $f23$, and $f25$. BSO and DE have the second and third ranks, respectively. WOA-BAT outperforms well in 13 functions, while BSO is well in 8 functions. Performance of DE is sufficient in 3 functions, which are $f4$, $f5$, and $f6$.

However, in terms of standard deviation, the ABC result is the best in 8 functions. GA has the worst results in all functions and does not perform well compared to other algorithms. DE is the second worse algorithm.

Table 13 is created in order to obtain the ranking result of the optimization algorithms from Table 12. As a result, Table 13 illustrates that WOA-BAT has the best ranking among the five optimization algorithms. Overall ranking WOA-BAT is 1.6. However, BSO has 2.6. Accordingly, the difference between WOA-BAT and BSO is 1, so it can be said that the difference is significant. WOA-BAT and BSO have

Table 8: Summary of the basic CEC2019 functions.

| No. | Functions | $F_i^* = F_i(x^*)$ | $D$ | Search range |
|-----|-----------|-------------------|-----|--------------|
| 1 | Storn's Chebyshev polynomial fitting problem | 1 | 9 | $[-8192, 8192]$ |
| 2 | Inverse Hilbert matrix problem | 1 | 16 | $[-16384, 16384]$ |
| 3 | Lennard-Jones minimum energy cluster | 1 | 18 | $[-4, 4]$ |
| 4 | Rastrigin's function | 1 | 10 | $[-100, 100]$ |
| 5 | Griewangk function | 1 | 10 | $[-100, 100]$ |
| 6 | Weierstrass function | 1 | 10 | $[-100, 100]$ |
| 7 | Modified Schwefel's function | 1 | 10 | $[-100, 100]$ |
| 8 | Expand Schaffer's F6 function | 1 | 10 | $[-100, 100]$ |
| 9 | Happy Cat function | 1 | 10 | $[-100, 100]$ |
| 10 | Ackley function | 1 | 10 | $[-100, 100]$ |

Table 9: Comparison result of WOA-BAT and WOA.

| Function | WOA | | WOA-BAT | |
|----------|-----|-----|---------|-----|
| | avg | std | avg | std |
| 1 | $\mathbf{1.2E-74}$ | $5.89E-74$ | $1.34E-06$ | $4.07E-07$ |
| 2 | $\mathbf{2.37E-51}$ | $8.82E-51$ | $0.0074$ | $0.0013$ |
| 3 | $50945$ | $13527.44$ | $\mathbf{0.000216}$ | $0.001126$ |
| 4 | $52.426$ | $24.97794$ | $\mathbf{0.001}$ | $7.1E-05$ |
| 5 | $28.02927$ | $0.452718$ | $\mathbf{11.1554}$ | $13.89905$ |
| 6 | $0.4356$ | $0.212037$ | $\mathbf{1.58E-06}$ | $7.15E-07$ |
| 7 | $0.0026$ | $0.002513$ | $\mathbf{0.0037}$ | $0.0076$ |
| 8 | $-10424$ | $1668.107$ | $\mathbf{-12214}$ | $1084.168$ |
| 9 | $\mathbf{1.89E-15}$ | $1.04E-14$ | $2.9852$ | $9.107663$ |
| 10 | $\mathbf{4.8E-15}$ | $2.35E-15$ | $0.1201$ | $0.652436$ |
| 11 | $0.011$ | $0.042629$ | $\mathbf{7.2E-08}$ | $3.4E-08$ |
| 12 | $0.02$ | $0.010083$ | $\mathbf{1.27E-08}$ | $5.89E-09$ |
| 13 | $0.5672$ | $0.296041$ | $\mathbf{2.1E-07}$ | $1.07E-07$ |
| 14 | $3.258$ | $3.214906$ | $\mathbf{0.998}$ | $4.52E-16$ |
| 15 | $0.000566$ | $0.000369$ | $\mathbf{0.000384}$ | $0.000354$ |
| 16 | $\mathbf{-1.0316}$ | $6.78E-16$ | $\mathbf{-1.0316}$ | $6.78E-16$ |
| 17 | $0.3979$ | $9.73E-06$ | $\mathbf{0.39789}$ | $1.69E-16$ |
| 18 | $\mathbf{3}$ | $6.15E-05$ | $7.5$ | $10.23432$ |
| 19 | $-3.856$ | $0.009536$ | $\mathbf{-3.8623}$ | $0.002004$ |
| 20 | $-3.225$ | $0.101675$ | $\mathbf{-3.2822}$ | $0.057156$ |
| 21 | $\mathbf{-8.746}$ | $2.324189$ | $-8.4675$ | $2.42464$ |
| 22 | $-7.6138$ | $2.858764$ | $\mathbf{-9.697}$ | $1.830619$ |
| 23 | $-6.7571$ | $3.587922$ | $\mathbf{-9.9988}$ | $1.640539$ |

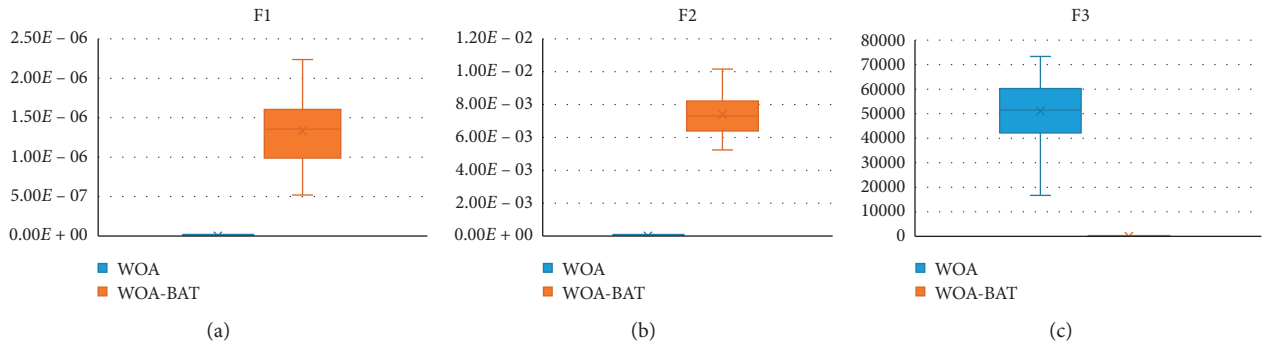

(a)                                     (b)                                     (c)

Figure 6: Continued.

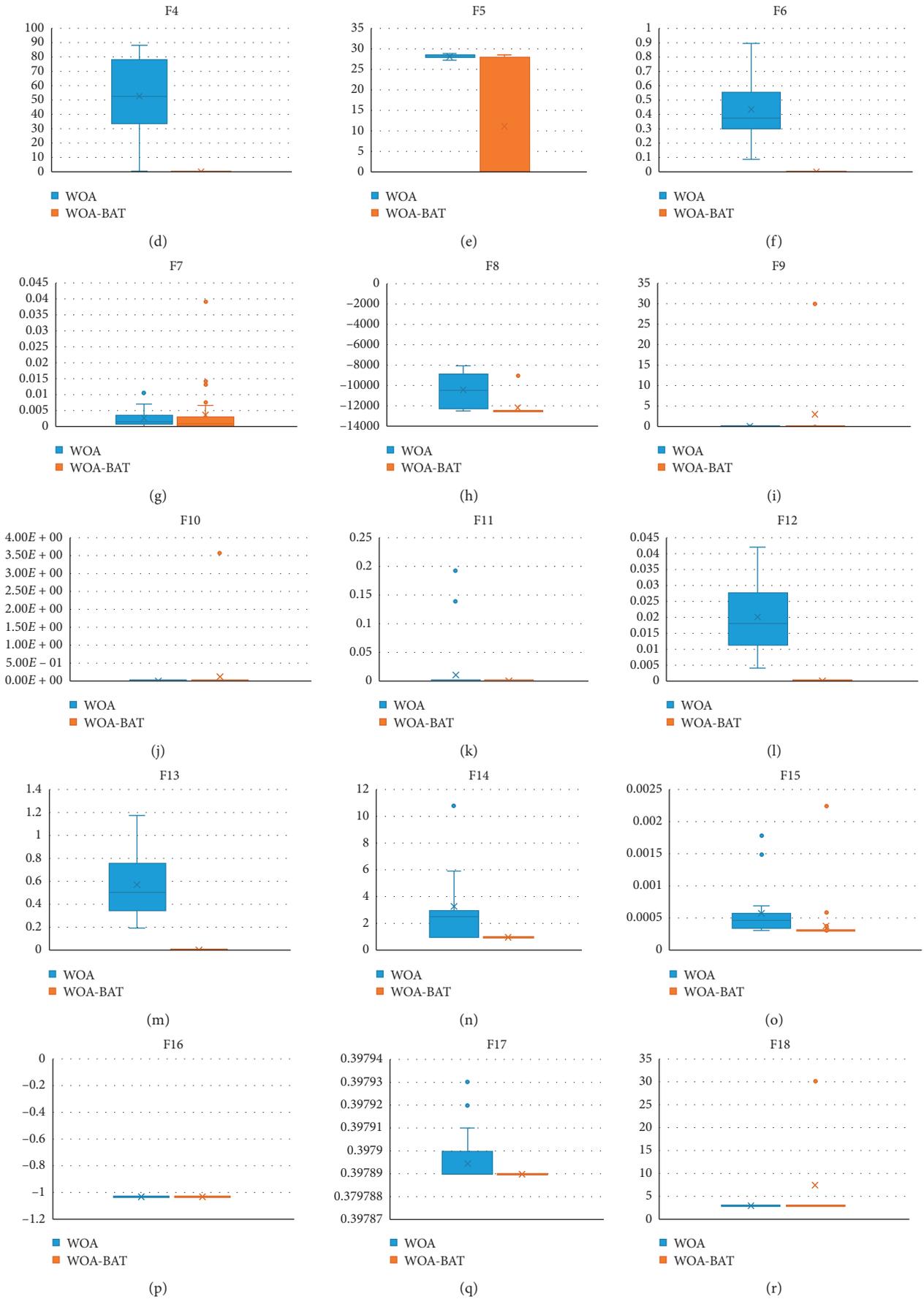

Figure 6: Continued.

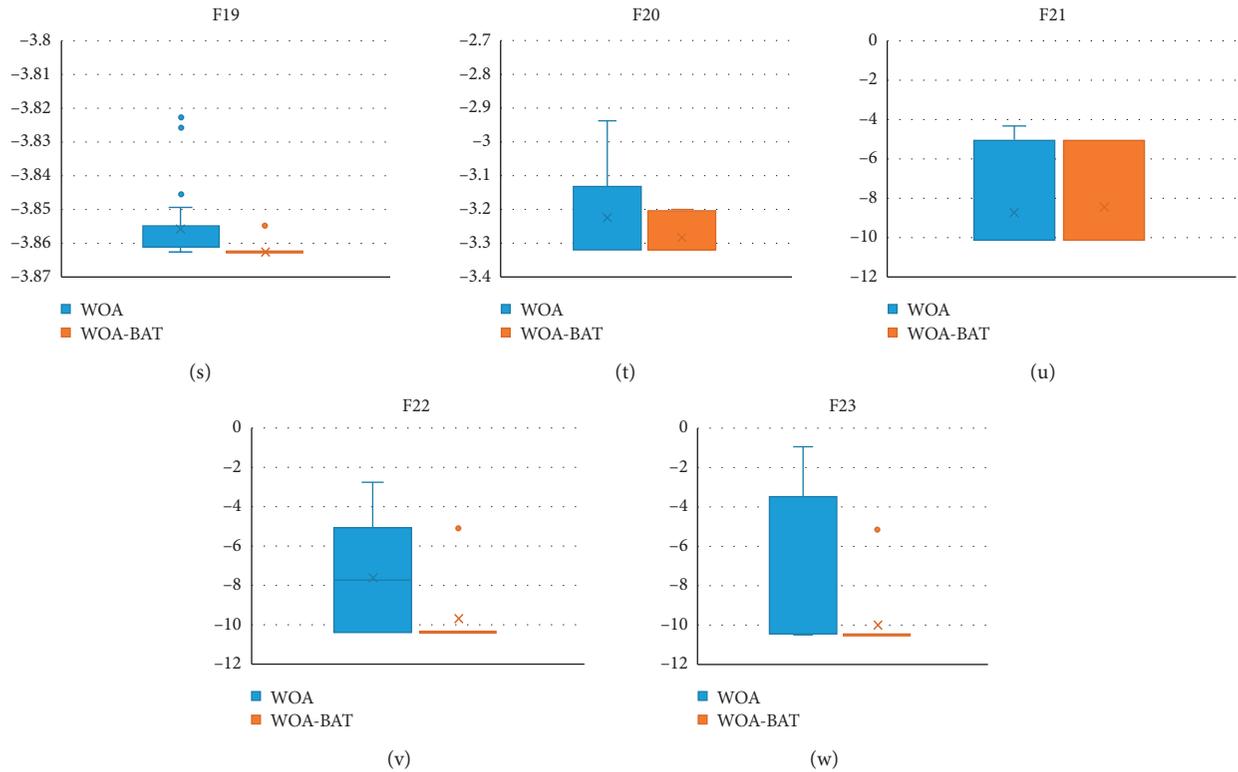

Figure 6: Comparison of average results of WOA-BAT and WOA.

Table 10: Comparison result of WOA-BAT and WOA on CEC2005.

| Function | WOA | | WOA-BAT | |
|---|---|---|---|---|
| | avg | std | avg | std |
| 1 | $8.83E + 00$ | $1.55E + 01$ | $\mathbf{3.94E + 00}$ | $5.47E + 00$ |
| 2 | $1.09E + 04$ | $3.96E + 03$ | $\mathbf{6.92E + 03}$ | $2.83E + 03$ |
| 3 | $3.02E + 06$ | $3.18E + 06$ | $\mathbf{1.33E + 06}$ | $1.72E + 06$ |
| 4 | $1.83E + 04$ | $6.56E + 03$ | $\mathbf{1.64E + 04}$ | $3.89E + 03$ |
| 5 | $\mathbf{2.87E + 03}$ | $2.89E + 03$ | $6.60E + 03$ | $3.68E + 03$ |
| 6 | $1.39E + 05$ | $2.42E + 05$ | $\mathbf{6.17E + 04}$ | $2.01E + 05$ |
| 7 | $\mathbf{1.27E + 03}$ | $2.98E - 01$ | $\mathbf{1.27E + 03}$ | $7.36E - 02$ |
| 8 | $\mathbf{2.03E + 01}$ | $9.86E - 02$ | $\mathbf{2.03E + 01}$ | $1.84E - 01$ |
| 9 | $4.22E + 01$ | $1.43E + 01$ | $\mathbf{3.46E + 01}$ | $1.10E + 01$ |
| 10 | $6.23E + 01$ | $2.22E + 01$ | $\mathbf{5.76E + 01}$ | $1.89E + 01$ |
| 11 | $\mathbf{8.87E + 00}$ | $1.22E + 00$ | $1.00E + 01$ | $1.75E + 00$ |
| 12 | $1.60E + 04$ | $1.38E + 04$ | $\mathbf{1.29E + 04}$ | $1.60E + 04$ |
| 13 | $4.42E + 00$ | $2.33E + 00$ | $\mathbf{4.17E + 00}$ | $2.44E + 00$ |
| 14 | $\mathbf{3.92E + 00}$ | $3.18E - 01$ | $4.01E + 00$ | $3.37E - 01$ |
| 15 | $\mathbf{2.19E + 01}$ | $3.88E + 01$ | $4.71E + 01$ | $4.38E + 01$ |
| 16 | $\mathbf{3.35E + 01}$ | $7.19E + 01$ | $5.51E + 01$ | $5.17E + 01$ |
| 17 | $\mathbf{2.29E + 01}$ | $3.78E + 01$ | $5.88E + 01$ | $3.62E + 01$ |
| 18 | $2.94E + 02$ | $1.41E + 02$ | $\mathbf{2.50E + 02}$ | $1.07E + 02$ |
| 19 | $2.77E + 02$ | $1.32E + 02$ | $\mathbf{2.67E + 02}$ | $9.22E + 01$ |
| 20 | $\mathbf{206.6743}$ | $170.0525$ | $250$ | $135.8244$ |
| 21 | $\mathbf{223.3337}$ | $212.8355$ | $267.9505$ | $122.4694$ |
| 22 | $330.573$ | $165.6103$ | $\mathbf{2.59E + 02}$ | $105.2167$ |
| 23 | $\mathbf{253.7131}$ | $242.8109$ | $294.9476$ | $126.9669$ |
| 24 | $\mathbf{199.7812}$ | $27.13892$ | $215.8896$ | $71.95231$ |
| 25 | $137.1858$ | $20.53592$ | $\mathbf{1.34E + 02}$ | $28.38144$ |

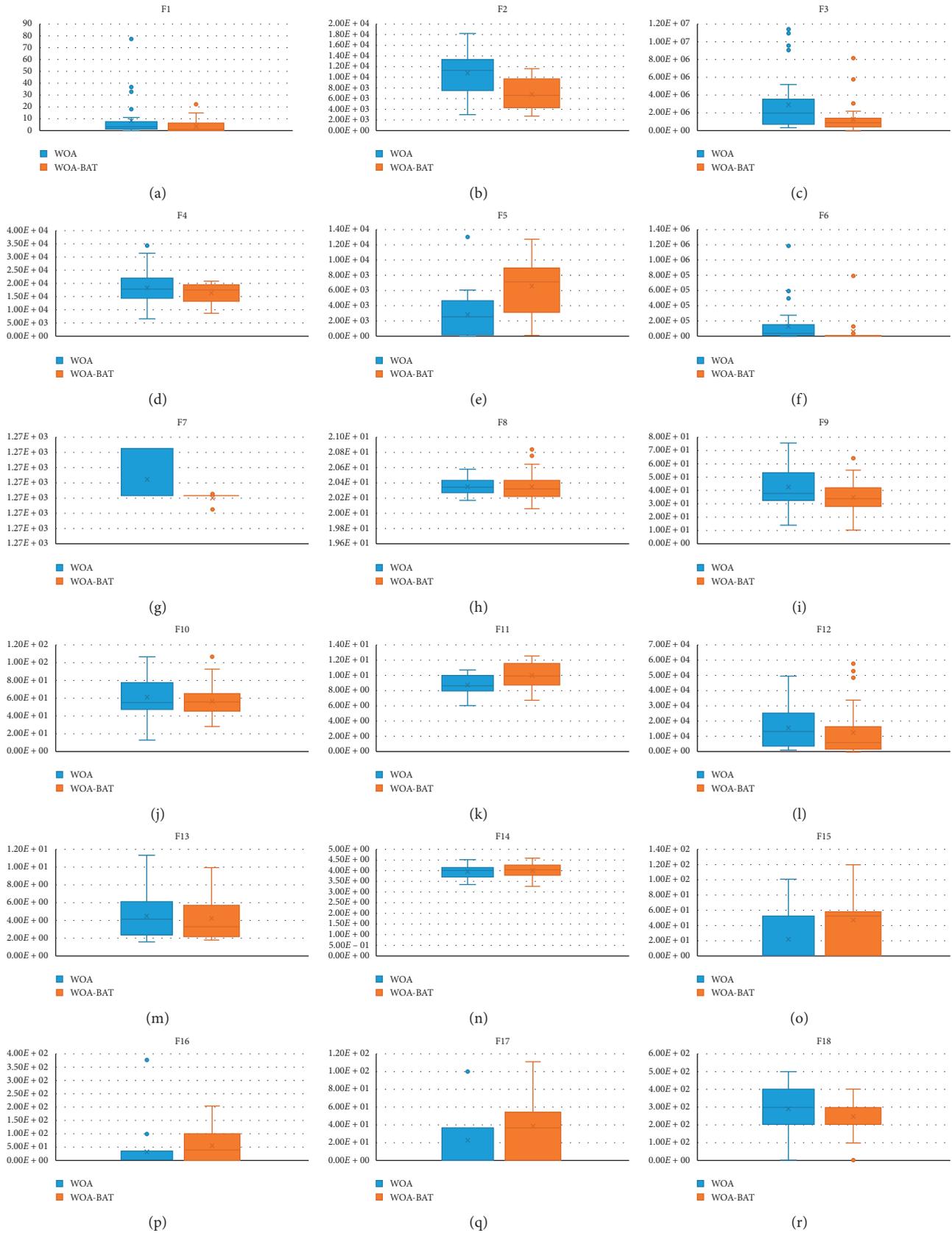

Figure 7: Continued.

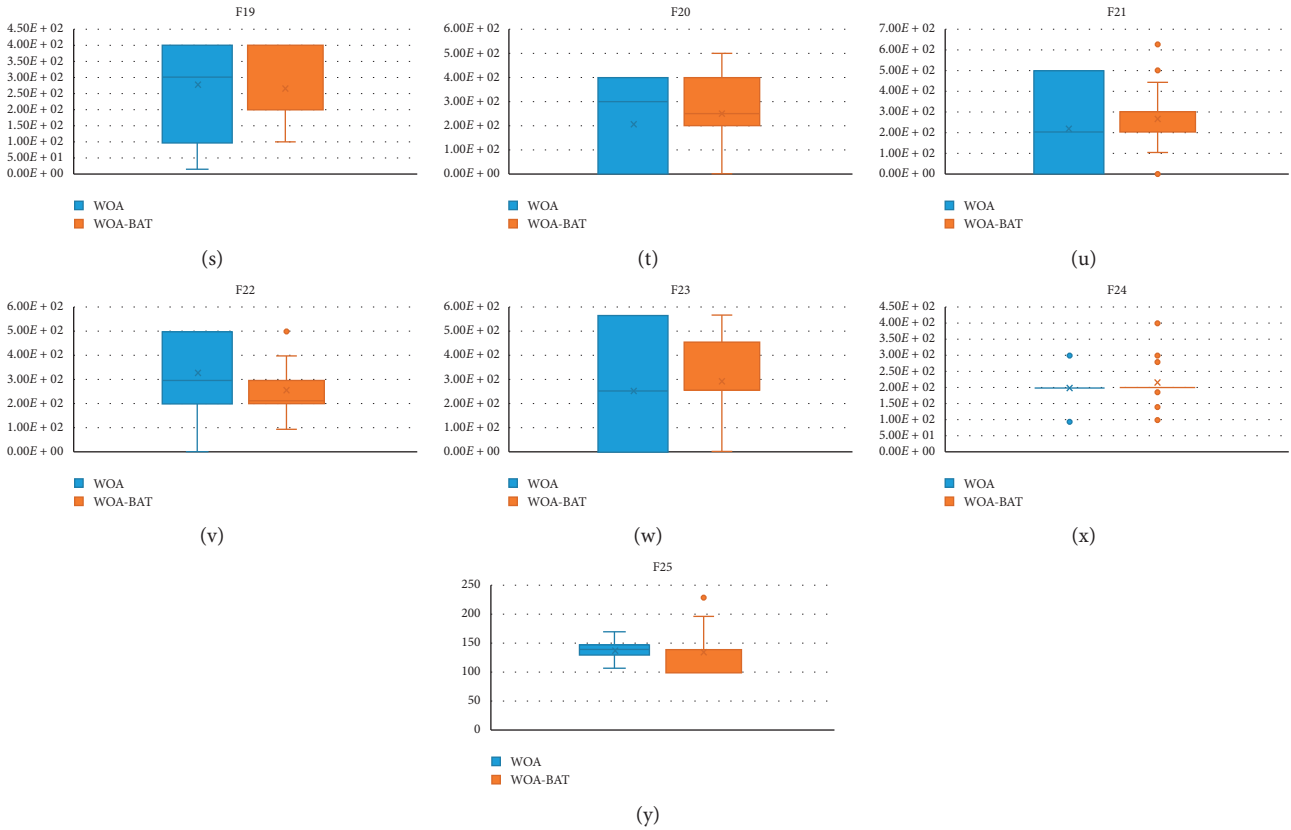

Figure 7: Comparison of average results of WOA-BAT and WOA CEC2005.

Table 11: Comparison results of WOA-BAT and WOA CEC2019.

| Function | WOA | | WOA-BAT | |
|---|---|---|---|---|
| | avg | std | avg | Std |
| 1 | $2.10E + 10$ | $3.57E + 10$ | $\mathbf{7.60E + 07}$ | $4.16E + 08$ |
| 2 | $1.84E + 01$ | $1.61E - 02$ | $\mathbf{1.75E + 01}$ | $1.21E - 01$ |
| 3 | $1.37E + 01$ | $7.23E - 15$ | $\mathbf{1.27E + 01}$ | $9.53E - 04$ |
| 4 | $\mathbf{3.48E + 02}$ | $1.72E + 02$ | $2.12E + 03$ | $1.01E + 03$ |
| 5 | $3.03E + 00$ | $4.86E - 01$ | $\mathbf{2.44E + 00}$ | $6.67E - 01$ |
| 6 | $\mathbf{1.03E + 01}$ | $1.39E + 00$ | $1.11E + 01$ | $1.55E + 00$ |
| 7 | $6.14E + 02$ | $2.98E + 02$ | $\mathbf{6.06E + 02}$ | $3.90E + 02$ |
| 8 | $6.03E + 00$ | $5.66E - 01$ | $\mathbf{5.72E + 00}$ | $7.18E - 01$ |
| 9 | $\mathbf{5.93E + 00}$ | $6.85E - 01$ | $2.28E + 01$ | $4.92E + 01$ |
| 10 | $2.13E + 01$ | $1.35E - 01$ | $\mathbf{2.12E + 01}$ | $2.26E - 01$ |

approximately the same ranking result for $f1$–$f12$. However, there is a significant difference between $f13$-$f14$ and $f15$–$f25$. WOA-BAT is better than BSO in $f15$–$f25$ by 1.9. Overall, it is believed that WOA-BAT has better ranking compare to GA, DE, ABC, and BSO.

## 8. Conclusion

In this study, WOA was explained in detail. WOA characteristics and its functionality were presented. In addition, the use of WOA was described in different areas, such as electrical and electronics engineering, automatic control system, civil engineering, fuel and energy, and medical engineering.

Furthermore, researchers have modified and hybridized WOA in order to overcome optimization problems in the above areas.

WOA was tested on 23 benchmark functions in order to determine the capability of exploitation, exploration, escaping from local minima, and convergence behavior. WOA had better performance of exploitation when it was tested on unimodal functions. It was also performed well in exploration on multimodal functions. In addition, testing WOA on composite functions can be viewed as the best way to stabilize between exploration and exploitation. Therefore, it can be said that WOA would increase convergence speed during iterations, while the majority of optimization algorithms (like PSO and GSA) do not have operators to consecrate a particular iteration to the exploration or the exploitation because they use only one format to update the position of search agents; thus, the probability of falling into local optima is more likely increased.

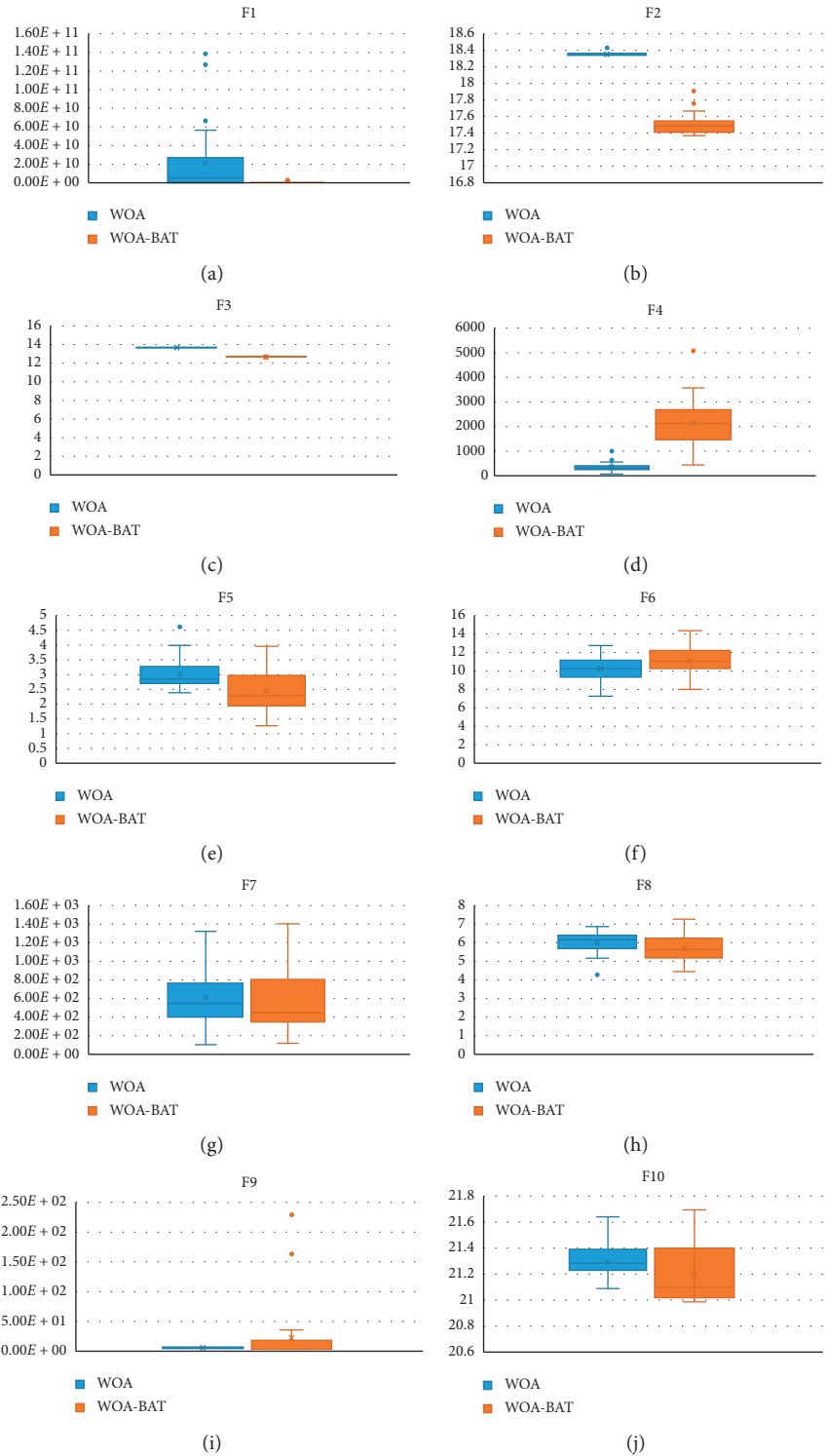

Figure 8: Comparison average result of WOA-BAT and WOA CEC2019.

It is safe to say that WOA achieves convergence speed and avoids local optima at the same time through iterations because of having two independent stages (exploration and exploitation). Both exploration and exploitation are done in each iteration.

Table 12: Comparison of WOA-BAT with GA, DE, ABC, and BSO.

| Function | GA | | DE | | ABC | | BSO | | WOA-BAT | |
|---|---|---|---|---|---|---|---|---|---|---|
| | avg | Std | avg | std | avg | std | avg | std | avg | std |
| 1 | $2.45E+03$ | $7.30E+02$ | $1.79E-04$ | $1.31E-04$ | $2.20E-02$ | $4.08E-02$ | $\mathbf{-4.50E+02}$ | $3.50E-14$ | $3.94E+00$ | $5.47E+00$ |
| 2 | $3.26E+04$ | $1.08E+04$ | $2.12E+02$ | $9.29E+01$ | $2.73E+04$ | $4.05E+03$ | $\mathbf{-4.48E+02}$ | $9.36E-01$ | $6.92E+03$ | $2.83E+03$ |
| 3 | $1.56E+08$ | $6.85E+07$ | $6.28E+06$ | $2.09E+06$ | $1.22E+08$ | $2.90E+07$ | $2.04E+06$ | $7.23E+05$ | $\mathbf{1.33E+06}$ | $1.72E+06$ |
| 4 | $1.30E+05$ | $6.17E+04$ | $\mathbf{1.15E+03}$ | $7.23E+02$ | $3.38E+04$ | $4.49E+04$ | $2.78E+04$ | $8.05E+03$ | $1.64E+04$ | $3.89E+03$ |
| 5 | $1.47E+04$ | $2.76E+03$ | $\mathbf{5.63E+02}$ | $2.84E+02$ | $8.30E+03$ | $8.00E+02$ | $4.70E+03$ | $1.22E+03$ | $6.60E+03$ | $3.68E+03$ |
| 6 | $6.71E+07$ | $3.87E+07$ | $\mathbf{3.94E+01}$ | $2.98E+01$ | $3.65E+05$ | $2.58E+05$ | $1.26E+03$ | $9.48E+02$ | $6.17E+04$ | $2.01E+05$ |
| 7 | $5.34E+03$ | $8.55E+01$ | $4.70E+03$ | $9.01E-11$ | $4.89E+03$ | $2.88E+01$ | $\mathbf{6.25E+02}$ | $3.25E+02$ | $1.27E+03$ | $7.36E-02$ |
| 8 | $2.10E+01$ | $6.64E-02$ | $2.10E+01$ | $7.75E-02$ | $2.10E+01$ | $6.86E-02$ | $\mathbf{-1.20E+02}$ | $9.90E-02$ | $2.03E+01$ | $1.84E-01$ |
| 9 | $7.86E+01$ | $1.36E+01$ | $1.46E+02$ | $2.87E+01$ | $2.10E+02$ | $1.35E+01$ | $\mathbf{-2.86E+02}$ | $1.27E+01$ | $3.46E+01$ | $1.10E+01$ |
| 10 | $3.39E+02$ | $4.09E+01$ | $2.15E+02$ | $1.13E+01$ | $2.46E+02$ | $9.04E+00$ | $\mathbf{-2.93E+02}$ | $8.79E+00$ | $5.76E+01$ | $1.89E+01$ |
| 11 | $3.56E+01$ | $2.71E+00$ | $4.04E+01$ | $1.35E+00$ | $4.05E+01$ | $1.37E+00$ | $1.10E+02$ | $2.51E+00$ | $\mathbf{1.00E+01}$ | $1.75E+00$ |
| 12 | $1.95E+05$ | $5.95E+04$ | $1.82E+04$ | $1.19E+04$ | $4.02E+05$ | $5.17E+04$ | $2.84E+04$ | $1.99E+04$ | $\mathbf{1.29E+04}$ | $1.60E+04$ |
| 13 | $1.49E+01$ | $2.53E+00$ | $1.79E+01$ | $1.49E+00$ | $2.31E+01$ | $1.45E+00$ | $\mathbf{-1.26E+02}$ | $1.05E+00$ | $4.17E+00$ | $2.44E+00$ |
| 14 | $1.34E+01$ | $2.28E-01$ | $1.37E+01$ | $1.32E-01$ | $1.36E+01$ | $1.34E-01$ | $\mathbf{-2.87E+02}$ | $3.78E-01$ | $4.01E+00$ | $3.37E-01$ |
| 15 | $5.47E+02$ | $6.41E+01$ | $2.70E+02$ | $9.66E+01$ | $3.06E+02$ | $5.76E+00$ | $5.43E+02$ | $7.94E+01$ | $\mathbf{4.71E+01}$ | $4.38E+01$ |
| 16 | $4.33E+02$ | $8.20E+01$ | $2.54E+02$ | $4.05E+01$ | $2.63E+02$ | $9.94E+00$ | $2.87E+02$ | $1.34E+02$ | $\mathbf{5.51E+01}$ | $5.17E+01$ |
| 17 | $8.34E+02$ | $2.25E+02$ | $2.81E+02$ | $4.62E+01$ | $2.86E+02$ | $1.72E+01$ | $3.10E+02$ | $1.57E+02$ | $\mathbf{3.88E+01}$ | $3.62E+01$ |
| 18 | $9.60E+02$ | $1.42E+01$ | $9.06E+02$ | $7.56E-01$ | $9.60E+02$ | $5.84E+00$ | $9.17E+02$ | $1.36E+00$ | $\mathbf{2.50E+02}$ | $1.07E+02$ |
| 19 | $9.57E+02$ | $1.62E+01$ | $9.06E+02$ | $8.12E-01$ | $9.63E+02$ | $7.72E+00$ | $9.16E+02$ | $1.07E+00$ | $\mathbf{2.67E+02}$ | $9.22E-01$ |
| 20 | $9.58E+02$ | $1.17E+01$ | $9.06E+02$ | $4.04E-01$ | $9.60E+02$ | $6.53E+00$ | $9.16E+02$ | $1.36E+00$ | $\mathbf{250}$ | $135.8244$ |
| 21 | $1.01E+03$ | $1.72E+02$ | $5.59E+02$ | $1.79E+02$ | $5.10E+02$ | $3.45E+02$ | $9.27E+02$ | $1.37E+02$ | $267.9505$ | $122.4694$ |
| 22 | $1.20E+03$ | $8.52E+01$ | $8.77E+02$ | $1.04E+01$ | $1.08E+03$ | $2.19E+01$ | $1.21E+03$ | $1.99E+02$ | $\mathbf{2.59E+02}$ | $105.2167$ |
| 23 | $1.01E+03$ | $1.71E+02$ | $5.91E+02$ | $1.72E+02$ | $5.49E+02$ | $2.56E+01$ | $9.48E+02$ | $1.38E+02$ | $\mathbf{294.9476}$ | $126.9669$ |
| 24 | $9.17E+02$ | $1.56E+02$ | $9.20E+02$ | $1.70E+02$ | $\mathbf{2.00E+02}$ | $3.48E-02$ | $4.67E+02$ | $6.23E+00$ | $215.8896$ | $71.95231$ |
| 25 | $1.79E+03$ | $3.92E+01$ | $1.64E+03$ | $3.33E+00$ | $1.51E+03$ | $8.75E+00$ | $1.88E+03$ | $4.44E+00$ | $\mathbf{1.34E+02}$ | $28.38144$ |

Table 13: Ranking of WOA-BAT optimization compared to GA, DE, ABC, and BSO.

| Functions | 1st | 2nd | 3rd | 4th | 5th | Rank | Subtotal | BSO |
|---|---|---|---|---|---|---|---|---|
| 1 | BSO | WOA-BAT | GA | DE | ABC | 2 | | |
| 2 | BSO | DE | WOA-BAT | ABC | GA | 3 | | |
| 3 | WOA-BAT | BSO | DE | ABC | GA | 1 | | |
| 4 | DE | WOA-BAT | BSO | ABC | GA | 2 | | |
| 5 | DE | BSO | WOA-BAT | ABC | GA | 3 | 11 | 9 |
| 6 | DE | BSO | WOA-BAT | ABC | GA | 3 | | |
| 7 | BSO | WOA-BAT | DE | ABC | GA | 2 | | |
| 8 | BSO | WOA-BAT | GA | DE | ABC | 2 | | |
| 9 | BSO | WOA-BAT | GA | DE | ABC | 2 | | |
| 10 | BSO | WOA-BAT | DE | ABC | GA | 2 | | |
| 11 | WOA-BAT | GA | DE | ABC | BSO | 1 | | |
| 12 | WOA-BAT | DE | BSO | GA | ABC | 1 | 13 | 14 |
| 13 | BSO | WOA-BAT | GA | DE | ABC | 2 | | |
| 14 | BSO | WOA-BAT | GA | ABC | DE | 2 | 4 | 2 |
| 15 | WOA-BAT | DE | ABC | BSO | GA | 1 | | |
| 16 | WOA-BAT | DE | ABC | BSO | GA | 1 | | |
| 17 | WOA-BAT | DE | ABC | BSO | GA | 1 | | |
| 18 | WOA-BAT | DE | BSO | GA | ABC | 1 | | |
| 19 | WOA-BAT | DE | BSO | GA | ABC | 1 | | |
| 20 | WOA-BAT | DE | BSO | GA | ABC | 1 | | |
| 21 | WOA-BAT | ABC | DE | BSO | GA | 1 | | |
| 22 | WOA-BAT | DE | ABC | GA | BSO | 1 | | |
| 23 | WOA-BAT | ABC | DE | BSO | GA | 1 | | |
| 24 | ABC | WOA-BAT | BSO | GA | DE | 2 | | |
| 25 | WOA-BAT | ABC | DE | GA | BSO | 1 | 12 | 42 |
| | | | **Total** | | | 40 | **Total** | 67 |
| | | | **Overall rank** | | | 40/25 = 1.6 | **Overall rank** | 67/25 = 2.6 |
| | | | **F1–F5** | | | 11/5 = 2.2 | **F1–F5** | 9/5 = 1.8 |
| | | | **F6–F12** | | | 13/7 = 1.8 | **F6–F12** | 14/7 = 2 |
| | | | **F13–F14** | | | 4/2 = 2 | **F13–F14** | 2/2 = 1 |
| | | | **F15–F25** | | | 12/11 = 1.9 | **F15–F25** | 42/11 = 3.8 |

It is obvious that WOA cannot solve every optimization problems. However, it is very competitive with other common optimization algorithms. Another limitation of WOA is that WOA has poor convergence speed while searching around the global optimum.

It is established that there are many types of WOA modifications and hybridizations. It is impossible to compare each new proposed WOA with all other types, and there are different benchmark functions, which can be used to test any new modifications. Therefore, it is believed that creating a platform for the researchers is essential in order to upload their program. After that, it will be easy to conduct and compare all the modifications and hybridizations and decide which one is the best.

WOA demonstrated high performance in solving many optimization problems. Regardless of all the results, WOA exhibited slow convergence speed due to finding the global optimum. As a result, the BAT algorithm is used to recover the exploration capability of WOA. Thus, the WOA-BAT algorithm presented to obtain better results in fewer iterations compared to WOA.

In this paper, WOA-BAT and WOA were tested on 25 functions from CEC2005. The results indicate that WOA-BAT performance is much better than WOA in 13 functions and have the same result in two functions. Also, WOA-BAT is tested on CEC2019 and compared with WOA. WOA-BAT has a lower average than WOA in 7 out of 10 functions.

WOA-BAT was evaluated against other competitive algorithms using CEC2005. The results showed that WOA-BAT has the first rank among GA, DE, ABC, and BSO.

There are several areas in WOA that can be further researched in the future. Therefore, the following areas might be interesting for researchers:

(1) Hybridization of WOA with other population metaheuristic algorithm, such as the ant-lion algorithm

(2) Investigation on the adaptive value, which is responsible for the exploration and exploitation ability of WOA-BAT

(3) Solving real-world problems in health care filed by hybridizing WOA-BAT with another optimization algorithm would be interested

(4) Hybridization of other optimization algorithms with WOA-BAT for cluster head selection for IoT

(5) It is recommended to use WOA-BAT to train other advanced types of machine learning techniques such as Capsule Net, LSTM, and CNN

(6) Applying WOA-BAT for constrained optimization problems

(7) Applying WOA-BAT for discrete optimization problems

(8) Solving different business applications by using the WOA-BAT algorithm

(9) Using WOA-BAT for feature selection in data mining

(10) Using WOA-BAT in the text mining field

## Conflicts of Interest

The authors declare that they have no conflict of interest.

## Acknowledgments

The authors wish to express their deep thanks to the University of Kurdistan Hewler (UKH) for providing funds for conducting this research study.

## Composition Comments

1. We have introduce the algorithm citations with in the text as per style, hence please check the placements of algorithm citations and change if it is necessary.

# Author(s) Name(s)

It is very important to confirm the author(s) last and first names in order to be displayed correctly on our website as well as in the indexing databases:

**Author 1**
Given Names: Hardi M.
Last Name: Mohammed

**Author 3**
Given Names: Tarik A.
Last Name: Rashid

**Author 2**
Given Names: Shahla U.
Last Name: Umar

It is also very important for each author to provide an ORCID (Open Researcher and Contributor ID). ORCID aims to solve the name ambiguity problem in scholarly communications by creating a registry of persistent unique identifiers for individual researchers.

To register an ORCID, please go to the Account Update page (http://mts.hindawi.com/update/) in our Manuscript Tracking System and after you have logged in click on the ORCID link at the top of the page. This link will take you to the ORCID website where you will be able to create an account for yourself. Once you have done so, your new ORCID will be saved in our Manuscript Tracking System automatically.